\documentclass[10pt,twocolumn]{IEEEtran}

\usepackage{amsmath}
\usepackage{amsfonts}
\usepackage{dsfont}
\usepackage{algorithm}
\usepackage{algorithmic}
\usepackage{graphicx}
\usepackage{subfigure}
\usepackage{url}
\usepackage{bbding}
\usepackage[english]{babel}
\usepackage{colortbl}
\definecolor{mygray}{gray}{0.8}

\begin{document}
\title{Towards Robust Pattern Recognition: A Review}

\author{Xu-Yao Zhang, Cheng-Lin Liu, Ching Y. Suen~\thanks{Xu-Yao Zhang, Cheng-Lin Liu, Ching Y. Suen. Towards Robust Pattern Recognition: A Review. Proceedings of the IEEE, vol. 108, no. 6, pp. 894-922, June 2020.}}

\maketitle
\begin{abstract}
The accuracies for many pattern recognition tasks have increased rapidly year-by-year, achieving or even outperforming human performance. From the perspective of accuracy, pattern recognition seems to be a nearly-solved problem. However, once launched in real applications, the high-accuracy pattern recognition systems may become unstable and unreliable, due to the lack of robustness in open and changing environments. In this paper, we present a comprehensive review of research towards \emph{robust pattern recognition} from the perspective of breaking three basic and implicit assumptions: \emph{closed-world assumption}, \emph{independent and identically distributed assumption}, and \emph{clean and big data assumption}, which form the foundation of most pattern recognition models. Actually, our brain is robust at learning concepts continually and incrementally, in complex, open and changing environments, with different contexts, modalities and tasks, by showing only a few examples, under weak or noisy supervision. These are the major differences between human intelligence and machine intelligence, which are closely related to the above three assumptions. After witnessing the significant progress in accuracy improvement nowadays, this review paper will enable us to analyze the shortcomings and limitations of current methods and identify future research directions for robust pattern recognition.
\end{abstract}


\section{Introduction}\label{sec:introduction}
In human intelligence, the ability of recognizing patterns is the most fundamental cognitive skill in the brain and serves as the building block for other high-level decision making, which is historically shown to be crucial for our survival and evolution in complex environments. On the other hand, for the purpose of machine intelligence, pattern recognition is also an essential goal for both machine learning and artificial intelligence, where the solving of many high-level intelligent problems relies heavily on the success of automatic and accurate pattern recognition.

During the past decades (see the survey papers in 1968~\cite{nagy1968state}, 1980~\cite{Fu1980pr} and 2000~\cite{jain2000review}), many exciting achievements in pattern recognition have been reported, and most successful methods are statistical approaches~\cite{vapnik1998statistical}, such as parametric and nonparametric Bayes decision rules~\cite{fukunaga1990introduction}, support vector machines~\cite{cortes1995svm}, boosting algorithms~\cite{freund1996experiments}, and so on. To guarantee high accuracy, these models are usually built on some well-designed hand-crafted features. In traditional approaches, the choice of the feature representation strongly influences the classification performance~\cite{jain2000review}. Since 2006~\cite{hinton2006science}, the end-to-end approach of deep learning~\cite{lecun2015nature}, which simultaneously learns the feature and classifier directly from the raw data, has become the new cutting-edge solution for many pattern recognition tasks.

The accuracies on many problems have been increased significantly and rapidly from time to time. For example, on the MNIST (10-class handwritten digit) dataset, with convolutional neural networks~\cite{lecun1998gradient}, it is easy to achieve more than 99\% accuracy without traditional hand-crafted features~\cite{liu2003handwritten}. On the more challenging task of 1000-class ImageNet large scale visual recognition, the accuracy was improved year-by-year, for example, AlexNet~\cite{krizhevsky2012alexnet} (2012, 84.7\%), GoogLeNet~\cite{szegedy2015googlenet} (2015, 93.33\%), ResNet~\cite{he2016resnet} (2016, 96.43\%), and so on. The newest accuracies~\cite{imagenet} have already surpassed human-level performance with large margins. Actually, this kind of accuracy improvement and record-breaking phenomena happen all the time for different pattern recognition tasks, such as face recognition~\cite{taigman2014deepface,sun2016face}, speech recognition~\cite{dahl2012speech,Saon2017speech}, handwriting recognition~\cite{Graves2009handwriting, Zhang2017benchmark}, and so on. It seems that: from the perspective of accuracy, pattern recognition has become a well-solved problem.

\begin{figure*}[!t]
\centering
\includegraphics[width=0.8\textwidth]{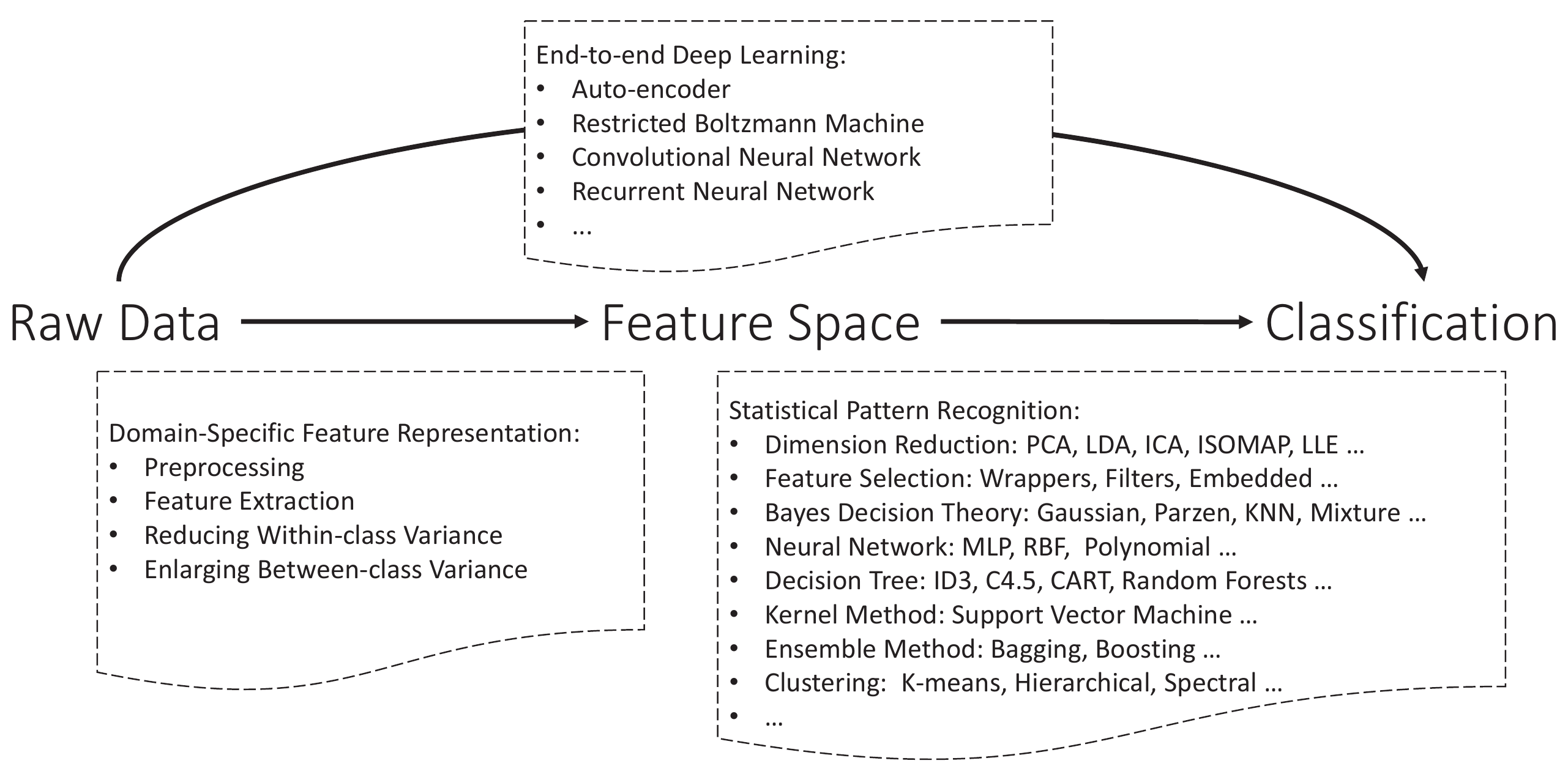}
\caption{A brief overview of pattern recognition methods.}\label{fig:pr_methods}
\end{figure*}

However, accuracy is only one aspect of measuring performance. When launching a high-accuracy pattern recognition system into real applications, many unsatisfying and unexpected results may still happen, making the system unstable in robustness, and the reason causing these problems is usually a mixture of different factors. For example, Nguyen et al.~\cite{nguyen2015fool} reveal that the state-of-the-art deep neural networks (DNNs) are easily fooled by assigning high confidence predictions for unrecognizable images, indicating that: although the accuracy of DNN is very high, it is not as robust as human vision when dealing with outliers. Moreover, as shown by~\cite{Goodfellow2015adversarial}, a small perturbation (particularly designed) on the input sample will cause a large perturbation (incorrect prediction) on the output of a pattern recognition system, leading to great adversarial risk when using such system in real applications with stringent safety requirement. Moreover, in traditional pattern recognition, the class set is usually assumed to be closed. However, in real world, the open set problem~\cite{Scheirer2013openSet} with dynamically changed class set is much more common. When re-contextualized into open set problems, many once solved tasks become significant challenging tasks again~\cite{Scheirer2014probabilityOpen}.

Another phenomenon causing significant performance drop for pattern recognition is the distribution mismatch. It has been shown that even a very small distribution shift can lead to a great performance drop for high accuracy pattern recognition systems~\cite{Recht2018cifar10}. Therefore, besides accuracy, the adaptability and transferability~\cite{Azizpour2016transferability} of a pattern recognition system become very important in real-life applications. Most pattern recognition systems only have single input and output, however, an effective strategy for improving robustness is to increase the diversity on both the input and the output of a system. Therefore, the multi-modal learning~\cite{Baltru2018multimodal} and multi-task learning~\cite{Caruana1997multitask} are also important issues for robust pattern recognition. In real world, the patterns seldom occur in isolation, but instead, usually appear with rich contextual information. Learning from the dependencies between patterns to improve the robustness of decision making is an important problem in pattern recognition~\cite{Haralick1983context}.

Most pattern recognition systems are actually data-hungry, and their high accuracies rely highly on both the quantity and quality of the training data. Any disturbances on the number or the labeling condition of the data will usually lead to great changes on the final performance. However, in real applications, it is usually difficult to collect large databases and produce accurate manual labeling. Therefore, the few-shot~\cite{Lake2015human} or even zero-shot~\cite{Lampert2014attribute} learning abilities of pattern recognition systems are of great value for real applications. On the other hand, in order to reduce the dependence on data quality, the pattern recognition system should be robust and learnable with noisy data~\cite{Frenay2014noise_survey}. Moreover, besides supervised training, other strategies like unsupervised~\cite{Bengio2013Review}, self-supervised~\cite{Doersch2017multi_task}, and semi-supervised~\cite{Miyato2018virtual} learning are also valuable for pattern recognition systems to learn from abundant unlabeled data and easily-obtained surrogate supervisory signals.

Based on the above observations, the problem of pattern recognition is far from being solved when considering different requirements in real applications. Besides accuracy, more attention should be paid on improving the robustness of pattern recognition. There are many previous works on robust pattern recognition (see Section~\ref{sec:robustness}), however, most of them are driven from a single view point of robustness. Currently, there is no clear definition on robust pattern recognition. In order to give a comprehensive understanding of robustness, and more importantly, reduce the gap between pattern recognition research and the requirements of real applications, in this paper we study and review different perspectives that are crucial for robust pattern recognition.

Actually, most pattern recognition models are derived from three implicit assumptions: \emph{closed-world assumption}, \emph{independent and identically distributed assumption}, and \emph{clean and big data assumption}. These assumptions are reasonable in a controlled laboratory environment and will simplify the complexity of the problem, therefore, they are fundamental for most pattern recognition models. However, in real-world, these assumptions are usually not satisfied, and in most cases, the performance of models built under these assumptions will deteriorate significantly. Therefore, to build robust systems for real environments, we should try to break these assumptions and develop new models and algorithms by reconsidering the essentials of pattern recognition.

In the rest of this paper, we first give a brief overview of pattern recognition methods in Section 2. After that, we define robustness for pattern recognition in Section 3. Then, we present detailed overviews on different attempts trying to break the three basic assumptions in Sections 4, 5 and 6, respectively. Lastly, we draw concluding remarks in Section 7. For readers who want to acquire some background knowledge before reading this paper, please refer to the appendix.

\section{A Brief Overview of Pattern Recognition Methods}
As shown in Fig.~\ref{fig:pr_methods}, pattern recognition methods can be divided into two main categories: two-stage and end-to-end. Most traditional methods are two-stage, i.e., with cascaded handcrafted feature representation and pattern classification.  The feature representation is to transform the raw data to a feature space with the property of within-class compactness and between-class separability. Preprocessing (like removing noise and normalizing data) is firstly applied to reduce within-class variance, while feature extraction further enlarges between-class variance, and this procedure is usually domain-specific. Actually, for solving new pattern recognition problems, the first thing is the design of feature representation, and a good feature will significantly reduce the burden on subsequent classifier learning. This kind of efforts can be found in different applications like iris recognition~\cite{Sun2009iris}, gait recognition~\cite{Wang2003gait}, action recognition~\cite{Wang2013action}, and so on.

After feature representation, the second stage is pattern classification, which is a much more general problem. Actually, classification is the main focus of many textbooks including Fukunaga~\cite{fukunaga1990introduction}, Duda et al.~\cite{Duda2012book}, Bishop~\cite{Bishop2006book}, and so on. This stage is also known as statistical pattern recognition~\cite{jain2000review}, where many different issues are considered from different perspectives. Firstly, dimensionality reduction~\cite{Yan2007graph} is widely adopted to derive a lower dimensional representation to facilitate subsequent classification task. Another approach of feature selection~\cite{Guyon2003selection} can be viewed as a discrete dimensionality reduction. After that, many classical classification models can be applied. The most fundamental one is the Bayes decision theory~\cite{fukunaga1990introduction} which integrates class-conditional density estimation with prior probability for maximum posterior probability classification. Artificial neural network is also widely used for pattern classification~\cite{Bishop1995neural}, including MLP (multilayer perceptron), RBF (radial basis function), polynomial networks, and so on. Decision tree based methods use a tree structure to represent the classification rule~\cite{Breiman2017cart}. Kernel methods~\cite{Lampert2009kernel, Muller2001kernel} have been widely applied to extend linear models to nonlinear ones by performing linear operations on higher or even infinite dimensional space transformed implicitly by a kernel mapping function, and the most representative method is SVM (support vector machine)~\cite{cortes1995svm}. Ensemble methods~\cite{Kuncheva2004combine} can further improve the performance by combining predictions from multiple complementary models. Clustering~\cite{Jain2010clustering} is widely used as an unsupervised strategy for pattern recognition.

In two-stage methods, we usually have multiple choices for both feature representation and classifier learning. It is hard to predict which combination will lead to the best performance, and in practice, different pattern recognition problems usually have different optimal configurations according to domain-specific experiences. Contrarily, deep learning~\cite{lecun2015nature} methods are end-to-end by learning the feature representation and classification jointly from the raw data. In this way, the learned features and classifiers are more cooperative toward the given task in a data-driven manner, which is more flexible and discriminative than two-stage methods.

Formerly, deep neural networks are usually layer-wise pre-trained by unsupervised models like auto-encoder~\cite{Vincent2008autoencoder} and restricted Boltzmann machine~\cite{hinton2006science}. Nowadays, deeper and deeper neural networks can be trained end-to-end due to many improved strategies such as better initialization~\cite{Glorot2010initialization}, activation~\cite{Maas2013relu}, optimization~\cite{Kingma2015adam}, normalization~\cite{Ioffe2015bn}, architecture~\cite{he2016resnet}, and so on. Due to shared-weights architecture and local connectivity characteristic, the convolutional neural network~\cite{lecun1998gradient} has been successfully used in many visual recognition tasks like image classification~\cite{krizhevsky2012alexnet}, detection~\cite{Ren2017faster_rcnn}, segmentation~\cite{Long2015segmentation}, and so on. Moreover, due to the ability of dealing with arbitrary-length sequences, the recurrent neural network has been widely used for sequence-based pattern recognition like speech recognition~\cite{Graves2006CTC}, scene text recognition~\cite{Shi2017CRNN}, and so on. Furthermore, the attention mechanism~\cite{Cho2015attention} can further improve deep learning performance by focusing on the most relevant information. Nowadays, deep learning has become the cutting-edge solution for numerous pattern recognition tasks.

Besides the broad class of \emph{statistical} pattern recognition approaches, \emph{structural} pattern recognition has been developed for exploiting and understanding the rich structural information in patterns~\cite{Bunke1990}. Unlike statistical feature representation, the structure of patterns is of variable dimensionality, and can be viewed as in non-Euclidean space. String matching and graph matching are basic problems in structural pattern recognition. To improve the learning ability of structural pattern recognition problems, kernel methods (with graph kernel~\cite{Bunke2011}), probabilistic graphical models~\cite{Koller2009}, and graph neural networks~\cite{Zhou2018} have been used. Overall, the research and application of structural pattern recognition is less popular than that of statistical methods.

\begin{table}[!t]
\renewcommand{\arraystretch}{1.3}
\caption{Representative studies with ``robust recognition'', ``robust classification'', or ``robustness'' in their titles.}
\label{table:definition}
\centering
\begin{tabular}{|c|c|l|c|}
\hline
Year &Ref. &Definition of Robustness &Type \\ \hline
1992 &\cite{Hampshire1992robust} &Heterogeneous sources &II\\ \hline
1996 &\cite{Kharin1996robustness} &Small-sample effects, distortion of samples &III \\ \hline
1996 &\cite{Mammone1996robust} &Environmental differences &II \\ \hline
1999 &\cite{Chibelushi1999robust} &Train/test condition mismatch &II\\ \hline
2001 &\cite{Provost2001robust} &Imprecise and changing environments &II \\ \hline
2003 &\cite{Miller2003mixture} &New class discovery, outlier rejection &I \\ \hline
2007 &\cite{Serre2007robust} &Clutter, learn from a few examples &III \\ \hline
2011 &\cite{He2011robust} &Noise corruption and occlusion, outliers &III \\ \hline
2011 &\cite{Elhamifar2011robust} &Small number of training data &III \\ \hline
2014 &\cite{Liu2014robust} &Distribution difference &II \\ \hline
2017 &\cite{Carlini2017robust} &Adversarial attack &I \\ \hline
2017 &\cite{Fawzi2017robust} &Adversarial, random noise &I \\ \hline
2018 &\cite{Xu2018robust} &Outlier, feature noise, label noise &III \\ \hline
\end{tabular}
\end{table}

\begin{figure*}[!t]
\centering
\includegraphics[width=0.8\textwidth]{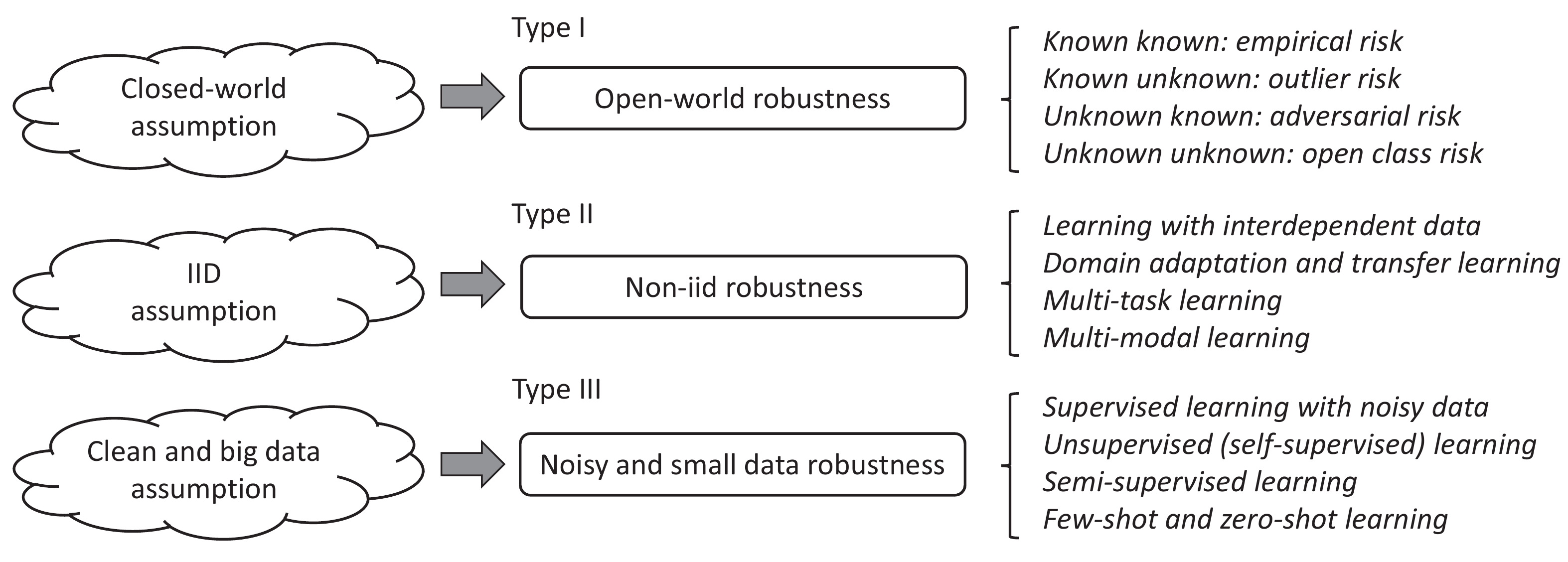}
\caption{Basic assumptions and robustness issues in pattern recognition.}\label{fig:robustness}
\end{figure*}

\section{Robustness in Pattern Recognition}\label{sec:robustness}
To build a pattern recognition system, there should be some training samples $\mathcal{D}_{\text{train}} = \{\boldsymbol{x}_i, y_i\}_{i=1}^n$ and test samples $\mathcal{D}_{\text{test}} = \{\hat{\boldsymbol{x}}_i, \hat{y}_i\}_{i=1}^{\hat{n}}$ where $\boldsymbol{x}$ is the observed pattern and $y$ is the corresponding label (the hat on the symbol is used to differentiate training and test samples). The purpose of pattern recognition is to learn the joint distribution $p(\boldsymbol{x}, y)$ or conditional distribution $p(y|\boldsymbol{x})$ from the training set $\mathcal{D}_{\text{train}}$ and then evaluate the learned model on a different test set $\mathcal{D}_{\text{test}}$. During this process, there are usually some basic assumptions.

\emph{Assumption I: Closed-world assumption.} The output space is assumed to be composed of a fixed number (e.g., $k$) of classes which are pre-defined as a prior $\Omega=\{1,2,\ldots,k\}$, and all samples are assumed to come from these classes $y_i \in \Omega$, $\hat{y}_i \in \Omega$, $\forall i$. Under this assumption, we can clearly and easily define the decision boundaries since the whole space is partitioned into $k$ regions. However, in real world applications, this assumption does not always hold, and there is an open space $\mathcal{O}$ much larger than $\Omega$. The samples in $\mathcal{O}$ can be outliers not belonging to any classes, unknown samples from some new classes not shown in the training set, or even adversarial samples from the confusing area. In these cases, the pattern recognition system will produce over-confident wrong predictions, because in its opinion there are only $k$ options and the winning class is highly reliable.

\emph{Assumption II: IID assumption}. The samples are assumed to be independent considering the joint distribution of observations and labels $p(\boldsymbol{x}_1, y_1, \boldsymbol{x}_2, y_2, \ldots, \boldsymbol{x}_n, y_n) = p(\boldsymbol{x}_1, y_1)p(\boldsymbol{x}_2, y_2) \ldots p(\boldsymbol{x}_n, y_n)$ or the marginal distribution of the observations $p(\boldsymbol{x}_1, \boldsymbol{x}_2,\ldots, \boldsymbol{x}_n) = p(\boldsymbol{x}_1)p(\boldsymbol{x}_2) \ldots p(\boldsymbol{x}_n)$, while the training and test data are assumed to be identically distributed $p(\boldsymbol{x}) \approx p(\hat{\boldsymbol{x}})$ and $p(\boldsymbol{x},y) \approx p(\hat{\boldsymbol{x}}, \hat{y})$. Under the independent assumption, we can then define the empirical loss as the summation of individual losses, e.g., $-\log p(\boldsymbol{x}_1, \boldsymbol{x}_2, \ldots, \boldsymbol{x}_n) = \sum_{i=1}^n -\log p(\boldsymbol{x}_i) $. Under the identical distribution assumption, we can therefore hope that minimizing the training error on $\mathcal{D}_{\text{train}}$ will yield good generalization performance on $\mathcal{D}_{\text{test}}$. However, in real world, the IID assumption is often violated: the data collected from multiple sources or conditions can not be simply viewed as independent, and moreover, a small mismatch between the training and test environments will cause significant performance degradation.

\emph{Assumption III: Clean and big data assumption}. The training data are assumed to be well labeled, and the volume of data is assumed to be large enough for covering different variations. Under this assumption, the only requirement is the capacity of the model, and supervised learning can be used to achieve good generalization performance. However, in real world applications, it is hard to collect a large number of training samples and also impossible to label all of them perfectly. How to effectively build pattern recognition systems from noisy data under small sample size (labeled or unlabeled) is a fundamental difference between machine intelligence and human intelligence.

As long as these assumptions remain stable, we can count on a reliable system to do its job time after time. However, when the assumptions no longer hold and the conditions start drifting, we want the system to keep its performance and be insensitive to these variations, this is called the \emph{robustness} for a pattern recognition system. Actually, in the literature, there are already a lot of research on \emph{robust pattern recognition} giving different definitions for robustness from diverse perspectives. We show some representative definitions in Table~\ref{table:definition}, and actually they can also be partitioned into three types corresponding to the above three assumptions. Usually, the lack of robustness in these studies is caused by the dissatisfaction of the assumptions. On the other hand, there are many other research works in the literature focusing on breaking the above assumptions but not using the terminology of \emph{robust pattern recognition}, which are also of great values to this field. Therefore, to better understand the current state and identify directions for future research, this paper surveys recent advances in robust pattern recognition and presents them in a common taxonomy according to the three assumptions.

The main contents are organized as shown in Fig.~\ref{fig:robustness}. Under each assumption, taxonomic sub-classification is presented to partition the content into four sub-topics, resulting in totally twelve issues. A brief and comprehensive overview is presented for each of them accompanied by a discussion on current and future research directions. For some of the topics there already exist good review papers, however, we differ from them by focusing more on recent advances and the relations to robust pattern recognition. Although these topics are interrelated, the review and discussion on each of them are made to be as self-contained as possible. Readers can choose to start at anywhere according to their own interests. However, it should always be remembered that the purpose is to break the three basic assumptions and realize robust pattern recognition.

\begin{figure*}[!t]
\centering
\includegraphics[width=0.6\textwidth]{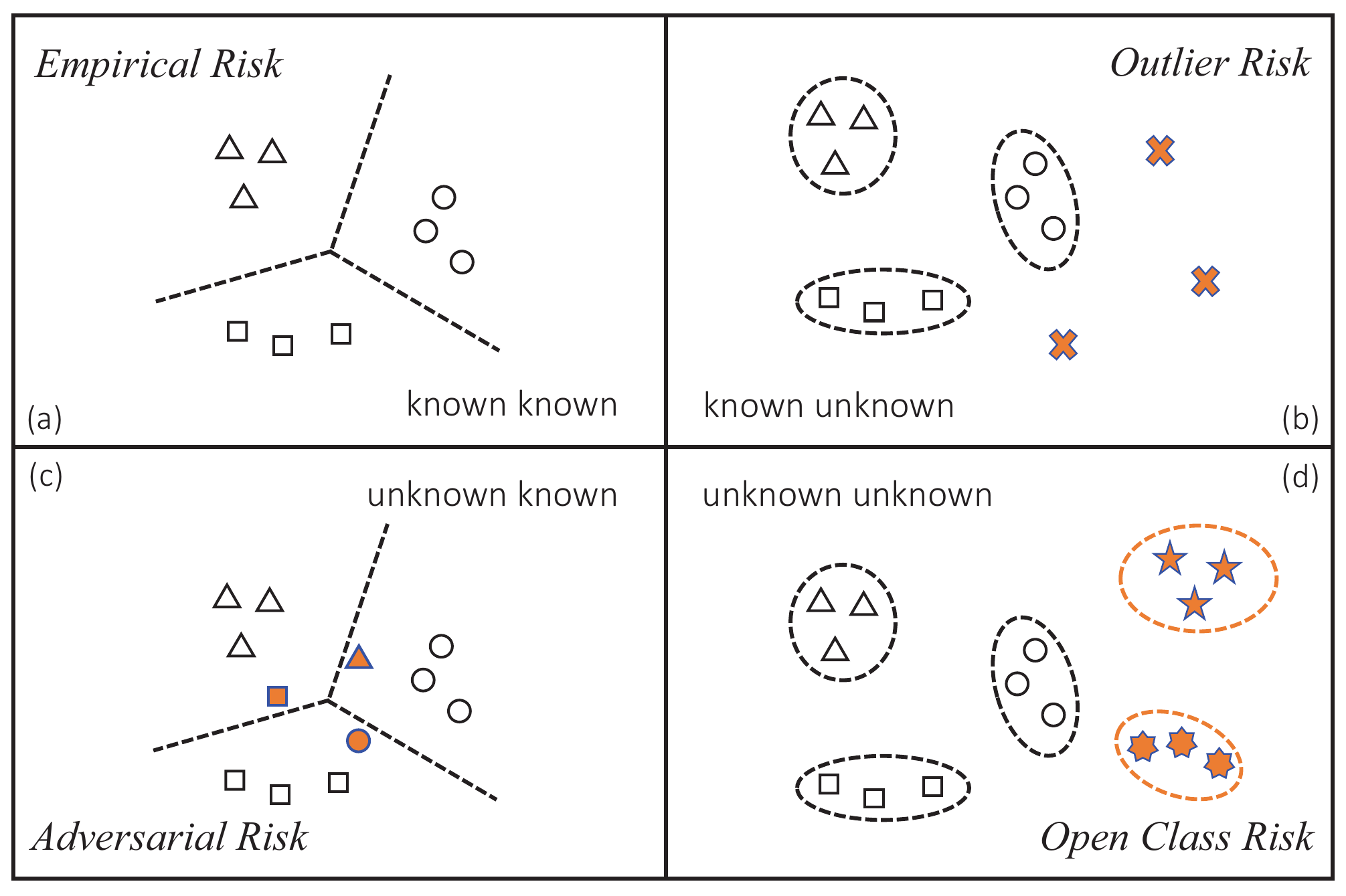}
\caption{An illustration on breaking closed-world assumption. The whole open space is partitioned into four parts: known known, known unknown, unknown known, and unknown unknown.}\label{fig:open_world}
\end{figure*}

\section{Breaking Closed-world Assumption}
Most pattern recognition methods are based on the closed-world assumption: \emph{although we only have finite observations of samples and categories, we still try to find a full partition of the whole space, this of course is unwise and improper}. For example, the support vector machine~\cite{cortes1995svm} seeks a hyperplane to partition the whole space into two half spaces under the principle of maximum margin. In deep neural networks~\cite{lecun2015nature}, the softmax layer partitions the whole space into a fixed number of classes and the summation of class probabilities is assumed to be one. These closed-world models will make overconfident errors on outliers and new category samples. Actually, there are massive unknown regions in pattern classification due to the finite set of training samples. To avoid making ridiculous mistakes, we must find methods to deal with these open and unseen spaces. In this paper, motivated by the ``\emph{known and unknown}'' statement in~\cite{Dietterich2017robustAI}, we summarize the approaches on breaking closed-world assumption into the following perspectives.

\subsection{Known Known: Empirical Risk}\label{sec:kk}
As shown in Fig.~\ref{fig:open_world}a, in closed-world recognition, we usually assume that we can observe some samples from some pre-defined categories, and we denote this case as ``\emph{known known}'' (things we know that we know). A straightforward strategy in this case is to minimize the \emph{empirical risk}, which is estimated as the misclassification rate on observed samples. However, since the number of samples is finite, minimizing empirical risk can not guarantee good generalization performance due to over-fitting. For example, it is common for the nearest neighbor decision rule~\cite{fukunaga1990introduction} to achieve perfect accuracy on training data but unsatisfactory performance on test data, and $k$-nearest neighbor is used to improve generalization with $k$ searched by cross-validation. In decision trees~\cite{Breiman2017cart}, over-fitting occurs when the tree is designed to perfectly fit all training samples, and pruning methods are applied to trim off unnecessary branches. Multilayer perceptron is able to approximate any decision boundary to arbitrary accuracy~\cite{Bishop1995neural}, and different tricks (like early stopping, weight decay) are used to avoid over-fitting.

The theoretical research of VC-dimension~\cite{vapnik1998statistical} suggests minimizing the \emph{structural risk} instead of empirical risk, by balancing the complexity of model against its success at fitting training data. Under this principle, the support vector machine~\cite{cortes1995svm} seeks a hyperplane that has the largest distance to the nearest training data point of any class, resulting in the large-margin regularization. From then on, many other regularization strategies have also been proposed, like sparsity~\cite{Zhang2010sparse}, low-rank~\cite{Dong2014lowRank}, manifold~\cite{Zhang2013manifold}, and so on. These regularization operations are usually combined with the empirical loss to build a better objective function. Other strategies (not integrated into the objective function) can also be viewed as implicit regularization, such as training with noise~\cite{Bishop1995noise}, regularized parameter estimation~\cite{Friedman1989RDA}, dropout~\cite{Srivastava2014dropout}, and so on. As a conclusion, a common strategy for ``\emph{known known}'' is empirical risk minimization with a well-defined regularization to improve generalization performance.

\subsection{Known Unknown: Outlier Risk}\label{sec:outlier}
As shown in Fig.~\ref{fig:open_world}b, besides known known, there is also ``\emph{known unknown}'' in the open space (we know there are some things we do not know). In open-world recognition, the things we do not know are often denoted as outliers. The simplest way to deal with outlier is extending the $k$-class problem to $k+1$ classes by adding a new class representing outliers. However, the drawback of this approach is that we need to collect outlier samples, and the distribution of outlier class is usually too complex to model. A more general case is that we do not have outlier samples, and the problem now becomes outlier detection~\cite{Hodge2004outlier}, anomaly detection~\cite{Chandola20009anomaly}, or novelty detection~\cite{Pimentel2014novelty}.

\subsubsection{Pattern rejection}
The first solution we should consider is to integrate some rejection strategies into the traditional pattern classifiers, and many such attempts can be found in the literature. For Bayes decision theory, Chow~\cite{chow1970reject} showed that the optimal rule (for rejecting ambiguous patterns) is to reject the pattern if the maximum of the a posteriori probabilities $\max_i p(i|\boldsymbol{x})$ is less than some threshold. Dubuisson and Masson~\cite{Dubuisson1993decision} proposed a modified rejection for the case $\sum_i p_i p(\boldsymbol{x}|i)$ being smaller than some threshold, which is suitable for rejecting outliers (not belonging to pre-defined classes). For many other classical models, the rejection option needs to be specifically designed according to the structure of classifier, like support vector machine~\cite{Grandvalet2009SVMreject}, nearest neighbor~\cite{Junior2017nearest}, sparse representation~\cite{Zhang2017sparseOpen}, multilayer perceptron~\cite{Karmakar2018neuralRejection}, and so on. Ensemble learning of multiple classifiers can also be used for rejection~\cite{Li2018overconfident}. It was shown that different classifier structures and learning algorithms affect the rejection performance significantly~\cite{Liu2002evaluation}.

\subsubsection{Softmax and extensions}
In many pattern recognition systems like deep neural networks, the \emph{softmax} function is widely used for classification:
\begin{equation}
p(i|\boldsymbol{x}) = \frac{e^{z_i (\boldsymbol{x})}}{ \sum_{j=1}^k  e^{z_j (\boldsymbol{x})}} \in [0,1], \ \ \ y = \mathop{\arg}\max_{i=1}^k p(i|{\boldsymbol{x}}),
\end{equation}
where $z_i (\boldsymbol{x})$ is the discriminant function for class $i$ and $y$ is the predicted class for $\boldsymbol{x}$. Let $p_1 = p(y|\boldsymbol{x})$ and $p_2 = \max_{i\neq y} p(i|\boldsymbol{x})$ be the top-1 and top-2 probabilities. To reject outliers, a straightforward strategy~\cite{Liu2002evaluation} is to set some thresholds on $p_1$ (confidence on predicted class) or $\triangle = p_1 - p_2$ (ambiguity between the top two classes), and the sample should be rejected if either $p_1$ or $\triangle$ is below some threshold. Actually, this kind of operation rejects uncertain predictions rather than unknown classes. Due to the closed-world property $\sum_{i=1}^k p(i|\boldsymbol{x}) = 1$, it can be easily fooled with outliers: a sample from a novel class (not predefined $k$ classes) may still have large values for both $p_1$ and $\triangle$. This means although the prediction is wrong, the classifier is still very confident (known as overconfident error~\cite{Li2018overconfident}), making the system hard to apply a threshold for rejection. A simple and straightforward modification is using the \emph{sigmoid} function
\begin{equation}
p(i|\boldsymbol{x}) = \frac{1}{1 + e^{-z_i (\boldsymbol{x})}} \in [0,1]
\end{equation}
to break the sum-to-one assumption and adopting the one-vs-all training~\cite{Liu2010_1vsAll, Shu2017open_text} to improve outlier rejection. In this case, for each class, the training samples from other classes are viewed as outliers, and a sample can be efficiently rejected if~\cite{Shu2017open_text} $\forall i: p(i|\boldsymbol{x}) < \text{threshold}$, since it does not belong to any known classes. Transforming sigmoid (binary) probabilities to multi-class probabilities satisfying $\sum_{i=1}^k p(i|\boldsymbol{x}) \le 1$ by the Dempster-Shafer theory of evidence~\cite{Liu2005confidence} can make outlier probability measurable as $1-\sum_{i=1}^k p(i|\boldsymbol{x})$.

To extend softmax for open set recognition, the \emph{openmax}~\cite{Bendale2016open_deep} fits a Weibull distribution on the distances between samples and class-means to give a parametric estimation of the probability for an input being an outlier with respect to each class. Another extension called \emph{generative openmax}~\cite{Ge2017gopenmax} employs generative adversarial network for novel category data synthesis to explicitly model the outlier class.

\subsubsection{One-class classification}
In the literature, another solution for outlier detection is the \emph{one-class classification}~\cite{Tax2001oneClass}, where all training samples are assumed to come from only one class. The support vector data description~\cite{Tax2004SVDD} uses a hyper-sphere with minimum volume to encompass as many training points as possible. The one-class SVM~\cite{Scholkopf2001one_class_svm} treats the origin in feature space as the representation for open space and maximizes the margin of training samples with respect to it using a kernel-based method. To use one-class models in multi-class recognition tasks, each class can be modeled with an individual one-class classifier, and then the outputs for different classes can be combined and normalized to grow a multi-class classifier with the reject option~\cite{Tax2008multi_class_reject}.

\subsubsection{Open space risk}
Recently, more and more attentions on this old and important issue are actually awakened due to the work of~\cite{Scheirer2013openSet}, which defined the \emph{open space risk} as:
\begin{equation}
R_\mathcal{O} (f) = \frac{\int_\mathcal{O} f(\boldsymbol{x}) d\boldsymbol{x}}{ \int_{S_o} f(\boldsymbol{x}) d\boldsymbol{x}},
\end{equation}
where $f$ is a measurable recognition function: $f(\boldsymbol{x}) > 0$ for recognition of the class of interest and $f(\boldsymbol{x}) = 0$ when it is not recognized. The $\mathcal{O}$ is the ``open space'' and $S_o$ is a ball that includes all of the known training samples as well as the open space $\mathcal{O}$. The $R_\mathcal{O} (f)$ is considered to be the relative measure of the open space compared to the whole space, and the challenge on using this theory lies on how to define $\mathcal{O}$ and get a computationally tractable open space risk term. In~\cite{Scheirer2013openSet}, the \emph{1-vs-set machine} is proposed as an extension of traditional SVM by using a slab defined by two parallel hyper-planes to define the open space. Similar idea has also been studied by~\cite{Cevikalp2017hyperplane} under open space hyper-plane classifiers. The work of~\cite{Scheirer2014probabilityOpen} further introduces a new model called \emph{compact abating probability} (CAP) by defining the open space $\mathcal{O}$ as the space sufficiently far from any known training sample:
\begin{equation}
\mathcal{O} = S_o - \bigcup_i B_r (\boldsymbol{x}_i),
\end{equation}
where $B_r (\boldsymbol{x}_i)$ is a closed ball of radius $r$ centered around training sample $\boldsymbol{x}_i$. A technique called Weibull-calibrated SVM has been proposed~\cite{Scheirer2014probabilityOpen} by combining CAP with statistical extreme value theory~\cite{Kotz2000EVT} for score calibration to improve multi-class open set recognition.

\subsubsection{Discussion}
Consider an expert pattern recognition system which can classify digits from 0 to 9 perfectly, when we feed an image of ``\emph{apple}'' into the system, it said ``\emph{this is a 6 and I am very confident}'', this will immediately change our feeling about this system from intelligent to foolish. The ability of \emph{learning to reject} is a major difference between closed-world and open-world recognition. Besides particular designed methods, theoretical analysis on this problem is particularly important, and although some studies have made good attempts on this direction, it is still worth further exploration. In many approaches, a threshold is usually used to distinguish normal and abnormal patterns, and different thresholds will lead to different tradeoffs between the adopted measurements (like precision and recall). Therefore, the choice of the threshold is usually task-dependent (different tasks will require different tradeoffs), and to evaluate the overall performance of a particular method, the threshold-independent metric~\cite{Hendrycks2017baseline} should be used, like the AUROC (area under receiver operating characteristic curve), AUPR (area under precision-recall curve), and so on.

\subsection{Unknown Known: Adversarial Risk}\label{sec:uk}
An intriguing phenomenon in open space is ``\emph{unknown known}'': things we think we know but it turns out we do not. Different from the \emph{known unknown} in Fig.~\ref{fig:open_world}b which denotes open space far away from training data and hence we know they are unknown, the \emph{unknown known} in Fig.~\ref{fig:open_world}c represents open space near decision boundaries where we are supposed to know but actually not, due to the limited number of training data not covering this space. Ambiguous prediction will happen for points close to the decision boundaries, for example, visually we think a sample is from one class but the system classifies it to another class. Since it is hard to sample such observation (low frequency in real world), the story is started by generating such samples to fool the system, which is known as \emph{adversarial examples}~\cite{Szegedy2013adversarial}.

\subsubsection{Generation of adversarial examples}
At the beginning, Szegedy et al.~\cite{Szegedy2013adversarial} show that by applying an imperceptible perturbation to an image, it is possible to arbitrarily change its prediction. Given any sample $\boldsymbol{x}$, an \emph{adversarial example} $\boldsymbol{x}'=\boldsymbol{x}+\eta$ can be found through constrained optimization. Since the perturbation $\eta$ is small, we can not find any obvious difference between $\boldsymbol{x}$ and $\boldsymbol{x}'$ visually, but their predicted labels are different, indicating the system is not robust: \emph{small perturbation on input causes large perturbation on output}. An efficient method to generate adversarial examples called \emph{fast gradient sign} is proposed by~\cite{Goodfellow2015adversarial}: let $\theta$ be parameters of a model, $\boldsymbol{x}$ and $y$ be input and ground truth, and $J(\theta, \boldsymbol{x}, y)$ be the cost used to train the model, the perturbation is then defined as:
\begin{equation}
\eta = \epsilon \cdot \text{sign} \left( \nabla_{\boldsymbol{x}} J(\theta, \boldsymbol{x}, y) \right),
\end{equation}
where $\epsilon>0$ is a step-parameter. The elements in $\eta$ correspond to the sign of the gradient of cost function with respect to input. Since $\eta$ is on the direction of gradient, moving $\boldsymbol{x}$ along $+\eta$ will increase $J$, and consequently, a large-enough $\epsilon$ will cause $\boldsymbol{x}'= \boldsymbol{x}+\eta$ to be misclassified. The \emph{iterative gradient sign}~\cite{Kurakin2016adversarial} is proposed as a refinement to fast gradient sign. After that, \emph{DeepFool} is proposed~\cite{Dezfooli2016deepfool} to search a minimal perturbation that is sufficient to change the label.

\subsubsection{Threat of adversarial examples}
As shown in~\cite{Carlini2017robust}, using adversarial examples as attacks will be great threats for many applications, such as self-driving cars, voice commands, robots, and so on. It has been shown in~\cite{Dezfooli2016deepfool} that a perturbation with 1/1000 magnitude as the original image is sufficient to fool state-of-the-art deep neural networks. Moreover, many new attack methods are still being gradually proposed~\cite{Papernot2016adversarial, Carlini2017robust}. The work of~\cite{Su2017pixel} showed that it is even possible to fool the system by only modifying one pixel of natural images. Furthermore, the method of~\cite{Papernot2017black_box} is proposed to attack a system which is viewed as a black-box. More surprisingly, the existence of a single small image-agnostic perturbation (called universal perturbation) that fools state-of-the-art classifiers on most natural images is also found~\cite{Dezfooli2017universal}. All these attempts have posed significant challenges on the robustness of pattern recognition systems.

\subsubsection{Defense methods}
To deal with adversarial attacks, many defense methods have been proposed. A typical approach is to augment the training set with adversarial examples and then retrain the model on the augmented data set~\cite{Goodfellow2015adversarial,Szegedy2013adversarial}. The \emph{defensive distillation} is proposed~\cite{Papernot2016distillation} to smooth the model during training for making it less sensitive to adversarial samples. The \emph{defense-GAN}~\cite{Samangouei2018gan} is trained to model the distribution of unperturbed real data, and at inference time it finds a close output to a given sample which does not contain the adversarial changes. Besides making the system robust to adversarial examples, Metzen et al.~\cite{Metzen2017detect} show that adversarial perturbations can also be detected by augmenting the system with a \emph{detector} network which is trained on the binary classification task of distinguishing genuine samples from perturbed ones. The \emph{deep contractive network}~\cite{Gu2014robust} adopts an objective function
\begin{equation}
\min_{\theta} L(\boldsymbol{x}) + \sum_{i=1}^H \lambda_i \| \frac{\partial h_i}{\partial h_{i-1}}\|_2,
\end{equation}
where $L(\boldsymbol{x})$ is a standard loss function and $h_i$ is the hidden representation for layer $i$ in a deep neural network. Since adversarial examples can be produced using the gradient sign method~\cite{Goodfellow2015adversarial}, regularizing the smoothness of gradient is therefore helpful to avoid them.

\begin{figure}[!t]
\centering
\includegraphics[width=\columnwidth]{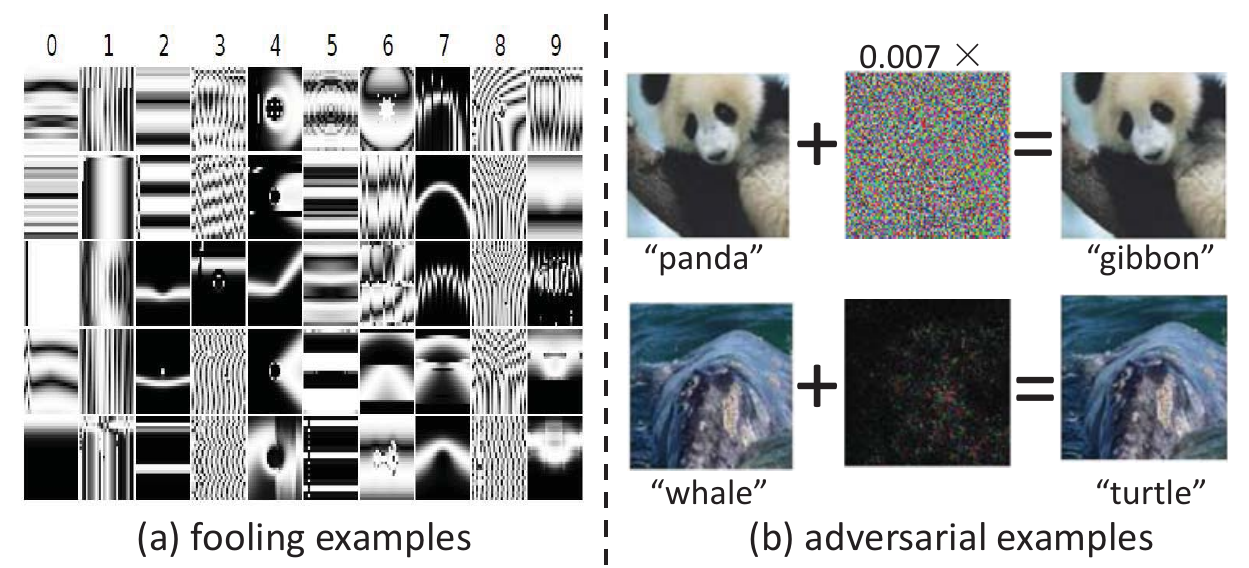}
\caption{(a): Fooling examples in~\cite{nguyen2015fool} which are classified to digits 0-9 with 99.99\% confidence. (b): Adversarial examples in~\cite{Goodfellow2015adversarial} and ~\cite{Dezfooli2016deepfool}.}\label{fig:adversarial}
\end{figure}

Since there are open spaces near the decision boundaries, \emph{data augmentation} can be used to fill them for improving robustness. The \emph{stability training}~\cite{Zheng2016stability} adds pixel-wise uncorrelated Gaussian noise for each input $\boldsymbol{x}$ to produce an augmented sample $\boldsymbol{x}'$, and then forces the outputs of the system on $\boldsymbol{x}$ and $\boldsymbol{x}'$ to be as close as possible, thus improving its robustness against small perturbations. The \emph{robust optimization}~\cite{Shaham2015robust} uses an alternating min-max procedure to increase local stability
\begin{equation}
\min_{\theta} \sum_{i=1}^n \max_{\boldsymbol{x}'_i \in \mathcal{U}_i} L(\boldsymbol{x}'_i, y_i),
\end{equation}
where $(\boldsymbol{x}_i,y_i)$ is a sample-label pair and $L$ is the loss function, the $\mathcal{U}_i$ is a ball around $\boldsymbol{x}_i$ with some radius. In the inside max procedure, the $\boldsymbol{x}'_i$ can be viewed as an augmented worst-case sample, and in the outside min problem, the loss on $(\boldsymbol{x}'_i, y_i)$ is minimized, thus making the system to be stable in a small neighborhood around every training point.

An interesting work of \emph{mixup}~\cite{Zhang2018mixup} proposes a data-agnostic method to produce augmented data points
\begin{equation}
\boldsymbol{x}' = \lambda \boldsymbol{x}_i + (1-\lambda) \boldsymbol{x}_j, \ \ \ \boldsymbol{y}' = \lambda \boldsymbol{y}_i + (1-\lambda) \boldsymbol{y}_j,
\end{equation}
where $\boldsymbol{x}_i, \boldsymbol{x}_j$ are two examples drawn at random from training data and $\boldsymbol{y}_i, \boldsymbol{y}_j$ are their corresponding one-hot label vectors.\footnote{Previously, we use $y$ to denote the label which is an integer from $\{1, 2, \ldots, k\}$. Here we use $\boldsymbol{y}$ (in bold) to represent a one-hot label vector: a vector (with length $k$) filled with $1$ at the index of the labeled class and with $0$ everywhere else.} The $\lambda \in [0,1]$ is a random parameter to produce the augmented sample $\boldsymbol{x}'$ and a new soft label $\boldsymbol{y}'$ (not one-hot anymore). This is based on the assumption that: linear interpolations of feature vectors should lead to linear interpolations of the associated labels. Although this approach is simple, it is very effective to produce augmented samples spreading not only within the same class but also between different classes, and therefore, the open space near decision boundaries is well handled with these augmented samples. Similar approach is also adopted in a work of \emph{between-class learning}~\cite{Tokozume2018betweenclass}. Although a mixture of two examples may not make sense for humans visually, it will make sense for machines as suggested by~\cite{Tokozume2018betweenclass}, and it is shown by~\cite{Zhang2018mixup} that this can not only improve the generalization performance but can also increase the robustness to adversarial examples.

\subsubsection{Discussion}
As shown in Fig.~\ref{fig:adversarial}a, another related concept is \emph{fooling examples}~\cite{nguyen2015fool} which are produced to be completely unrecognizable to human eyes but the pattern recognition system will still classify them into particular classes with high confidence. This is different from \emph{adversarial examples} shown in Fig.~\ref{fig:adversarial}b. Actually, the phenomenon of \emph{fooling examples} is the result of outliers with closed-world assumption which is discussed in Section~\ref{sec:outlier}. It is shown by~\cite{nguyen2015fool} that retraining of the system by viewing fooling examples as a newly added class is not sufficient to solve this problem, and a new batch of fooling images can be produced to fool the new system even after many retraining iterations. This is because there are massive open spaces for outliers and it is impossible to model them completely. Contrarily, augmenting training data with adversarial examples was shown to significantly increase the robustness even with only one extra epoch~\cite{Dezfooli2016deepfool}, this is because the open spaces near decision boundaries are limited and constrained, and therefore giving us possibility to model them. Since many people continue to propose new attack methods for producing adversarial examples, the research of novel defense strategy becomes particularly important to guarantee the safety of pattern recognition.

\subsection{Unknown Unknown: Open Class Risk}\label{sec:uu}
As shown in Fig.~\ref{fig:open_world}d, the last case in open space is ``\emph{unknown unknown}'': situations where a lot of unknown samples (out of this world) are grouping into different unknown (unseen) categories. In this case, we should not simply mark them as a single and large category of unknown (like Section~\ref{sec:outlier}), but also need to identify the newly emerged categories in a fine-grained manner. This is a common situation in real applications, where the datasets are dynamic and novel categories must be continuously detected and added, which is denoted as \emph{open world recognition} in~\cite{bendale2015open} and \emph{class-incremental learning} in~\cite{Rebuffi2017incremental}.

\subsubsection{Definition of the problem}
\begin{figure}[!t]
\centering
\includegraphics[width=\columnwidth]{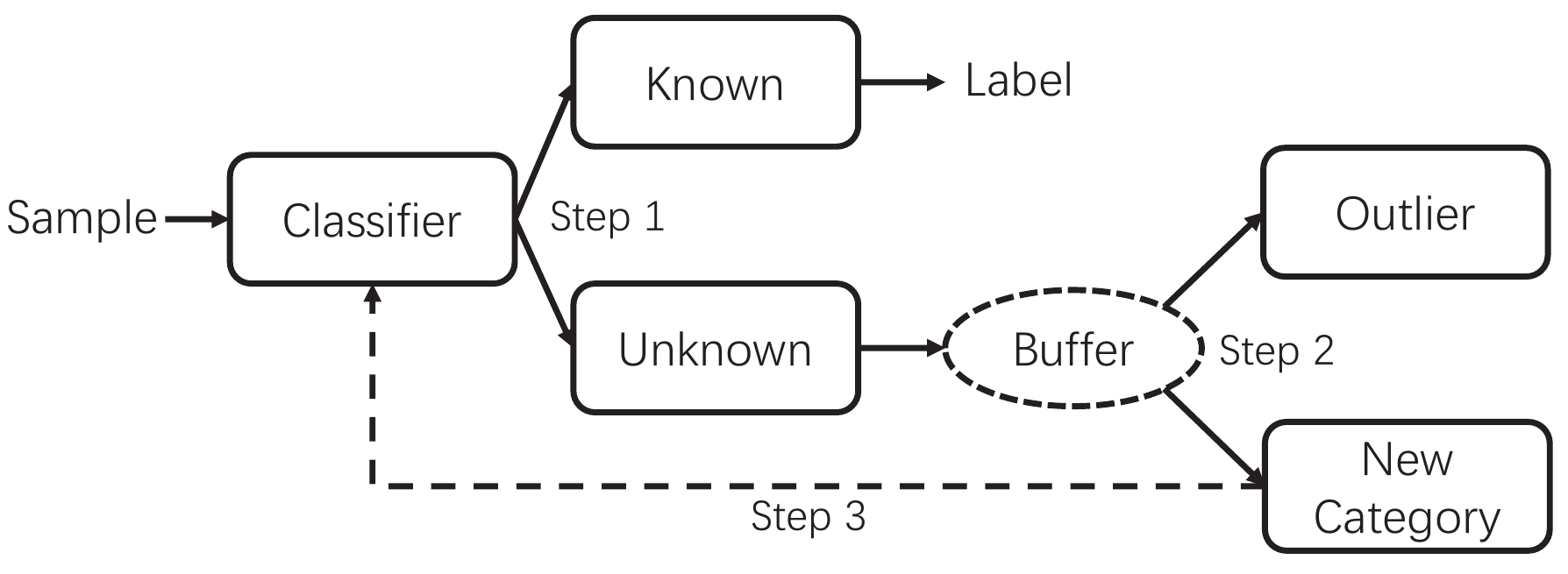}
\caption{Open-world recognition with class-incremental learning.}\label{fig:class_incremental}
\end{figure}

During continuous use of a pattern recognition system, abundant or even infinite test data will come in a streaming manner. As shown in Fig.~\ref{fig:class_incremental}, the open-world recognition process can be decomposed into three steps. The first step is detecting unknown samples and placing them in a buffer, which requires the system to reject samples from unseen classes and keep high accuracy for seen classes. The second step is labeling unknown samples in the buffer into new categories, which can be either finished by humans or automatically implemented. The last step is then updating the classifier with augmented categories and samples, which requires the classifier to be efficiently trainable in a class-incremental manner where different classes occur at different times. Step 1 has already been discussed in Section~\ref{sec:outlier}, therefore, this section focuses on steps 2 and 3.

\subsubsection{Labeling unknown samples}
A simple and accurate approach for step 2 is seeking help from human beings~\cite{bendale2015open}, either in a batch manner when the buffer size reaches some threshold, or immediately when the users encounter some strange outputs from the system and then try to give some feedback. Moreover, a strategy to make this process more efficient is using active learning~\cite{Huang2014active} to reduce the labeling cost by selecting the most valuable data to query their labels. On the contrary, a more challenging task is automatic new class discovery~\cite{Chang1991new_class} without human labeling. Unsupervised clustering~\cite{Masud2011novel_class} is an efficient and effective solution for finding new classes. A cluster can be seen as a new category if the number of samples falling in this cluster is large enough, and otherwise, it should only be viewed as an outlier and ignored. However, a difficulty for this approach is the model selection problem, i.e., how many novel classes are contained in the data? To deal with this, the clustering algorithms should have the ability of automatic model selection~\cite{Li2018cluster}.

\subsubsection{Class incremental learning}
The solution of step 3 requires us to rethink the relationship between \emph{discriminative} and \emph{generative} models. A pattern recognition system usually contains class-independent feature extraction $\phi(\boldsymbol{x})$ and class-specific decision function $g_i(\boldsymbol{x})$.\footnote{For example, in deep learning, the $\phi(\boldsymbol{x})$ is a multi-layer neural network, and $g_i(\boldsymbol{x})$ is usually a linear function on $\phi(\boldsymbol{x})$ like $g_i(\boldsymbol{x}) = w_i^{\top} \phi(\boldsymbol{x}) + b_i$.} The classification is then: $\boldsymbol{x} \in \arg \max_i g_i (\boldsymbol{x})$. In discriminative model, all the decision functions $g_i(\boldsymbol{x}), \forall i$ are trained \emph{jointly} (like hinge loss, softmax loss, and so on), which can be viewed as the competition between different classes to adjust decision boundaries. On the contrary, in generative model, $g_i(\boldsymbol{x})$ is usually used to model each class \emph{independently} (like negative log-likelihood loss for some distribution). Discriminative model usually has higher accuracy, however, since $g_1(\boldsymbol{x}),\ldots,g_k(\boldsymbol{x})$ are coupled in training, adding a new class $g_{k+1} (\boldsymbol{x})$ will affect others, requiring retraining of them with all data available. Contrarily, in generative model, class-incremental learning will become much simpler, since the training of $g_{k+1} (\boldsymbol{x})$ is independent of other classes. However, the drawback is that generative model usually leads to lower accuracy. Therefore, hybrid discriminative and generative models become necessary: $\phi(\boldsymbol{x})$ is discriminative while $g_i(\boldsymbol{x})$ is generative. Actually, many recent works are already using this principle.

\subsubsection{Prototype based approaches}
The nearest class mean (NCM)~\cite{Mensink2013ncm} can generalize to new classes at near-zero cost:
\begin{equation}
g_i (\boldsymbol{x}) = - \| \phi(\boldsymbol{x}) - \mu_i \|_2^2, \ \ \ \mu_i = \frac{1}{n_i} \sum_{j: y_j = i } \phi(\boldsymbol{x}_j),
\end{equation}
where $\mu_i$ is the mean of training samples (totally $n_i$) in $\phi(\boldsymbol{x})$ space for class $i$, and $g_i(\boldsymbol{x})$ is based on the Euclidean distance to class-mean.  Different criteria can be defined, such as softmax~\cite{Mensink2013ncm} or sigmoid~\cite{Rebuffi2017incremental}, to learn a discriminative $\phi(\boldsymbol{x})$ for high accuracy, and the updating of $\mu_i$ is always a generative mean calculation. When new class arrives, it is efficient to compute $\mu_{\text{new}}$ and augment the model with a new decision function $g_{\text{new}} (\boldsymbol{x})$, without affecting other classes.

The class-independent feature extraction $\phi(\boldsymbol{x})$ can be either a linear dimensionality reduction~\cite{Mensink2013ncm} or a nonlinear deep neural network~\cite{Guerriero2018deepncm}. Similar idea is also adopted in~\cite{Rebuffi2017incremental} where a lot of exemplar samples are selected dynamically to represent each class in CNN transformed space, and a nearest-mean-of-exemplars strategy is used for classification. Another work of \emph{prototypical network}~\cite{Snell2017prototypical} is also a NCM classifier in deep neural network transformed space. A more general analysis on representing each class as Gaussian is given in~\cite{Wan2018loss}. A work of \emph{convolutional prototype learning}~\cite{Yang2018prototype} uses automatically learned prototypes to represent each class by regularizing the deviation of prototypes from class-means. As shown in~\cite{Qi2018imprint}, when normalized to lie on a sphere, NCM is also equal to the traditional linear classifier in neural networks.

\subsubsection{Discussion}
Since it is hard to enumerate all categories at once, how to smoothly update the system to learn more and more concepts over time is therefore an important and challenging task. Although using class-means or prototypes to represent each class is a simple generative model, it is effective for class distribution modeling, because a powerful $\phi(\boldsymbol{x})$ (e.g., deep neural networks) can be learned to transform complex intra-class distributions into simplified Gaussian distributions, which will then be efficient and effective for class-incremental learning. Other classical classifiers can also be modified for class-incremental learning like random forest~\cite{Hu2018random_forest}, support vector machine~\cite{Kuzborskij2013incremental}, and so on. For class-incremental learning of deep neural networks, the newly learned classes may erase the knowledge of old classes, due to the joint updating of $\phi(\boldsymbol{x})$ and $g_i(\boldsymbol{x})$~\cite{Rebuffi2017incremental}, resulting in \emph{catastrophic forgetting}~\cite{Li2018forgetting}. A remedy is to review the historical data of old classes occasionally to prevent forgetting~\cite{He2018class_incremental}, and many other recent advances are gradually proposed on this topic like the evolving neural structure (learning to grow)~\cite{Li2019continual_structure}, dynamic generative memory (learning to remember)~\cite{Ostapenko2019remember}, and so on. More efficient class-incremental learning models that can handle forgetting problem effectively will be the focus of future research.

\section{Breaking IID Assumption}
Independent and identically distributed (IID) random variable is a fundamental assumption for most pattern recognition methods. However, in practical applications, this assumption is often violated. On the Dagstuhl seminar organized by Darrell et al.~\cite{Darrell2015IIDSeminar} in 2015, it was the agreement of all participants that learning with interdependent and non-identically distributed data should be the focus of future research. Moreover, it was shown by~\cite{Recht2018cifar10} that even a very small mismatch between training and test distributions will make state-of-the-art models dropping their performance significantly.

In pattern recognition, a labeled sample $(\boldsymbol{x},y)$ is usually assumed to come from a \emph{feature space} $\boldsymbol{x} \in \mathcal{X}$ and a \emph{label space} $y \in \mathcal{Y}$. A specific combination of feature space and label space can be viewed as an \emph{environment} $\mathcal{E}=\mathcal{X} \times \mathcal{Y}$, where a \emph{learner} $p(y|\boldsymbol{x})$ is defined and performed. The learner should be adjustable when the environment starts to change $\mathcal{E} \neq \mathcal{E}'$, which can be summarized in four cases:
\begin{itemize}
\item $\mathcal{X} = \mathcal{X}'$ and $\mathcal{Y} = \mathcal{Y}'$. This is the most-widely considered case, where the feature spaces and label spaces are identical, and the environmental change comes from the conditional distribution $p(y|\boldsymbol{x}) \neq p(y'|\boldsymbol{x}')$.
\item $\mathcal{X} = \mathcal{X}'$ and $\mathcal{Y} \neq \mathcal{Y}'$. The feature spaces are identical but the label spaces are different, for example, cross-class transfer learning and multi-task learning.
\item $\mathcal{X} \neq \mathcal{X}'$ and $\mathcal{Y} = \mathcal{Y}'$. The feature spaces are different while the label spaces are identical, which happens often in multi-modal learning.
\item $\mathcal{X} \neq \mathcal{X}'$ and $\mathcal{Y} \neq \mathcal{Y}'$. Both the feature spaces and label spaces are different, which is the most difficult situation, and can be viewed as multi-modal multi-task learning.
\end{itemize}

\begin{figure}[!t]
\centering
\includegraphics[width=0.95\columnwidth]{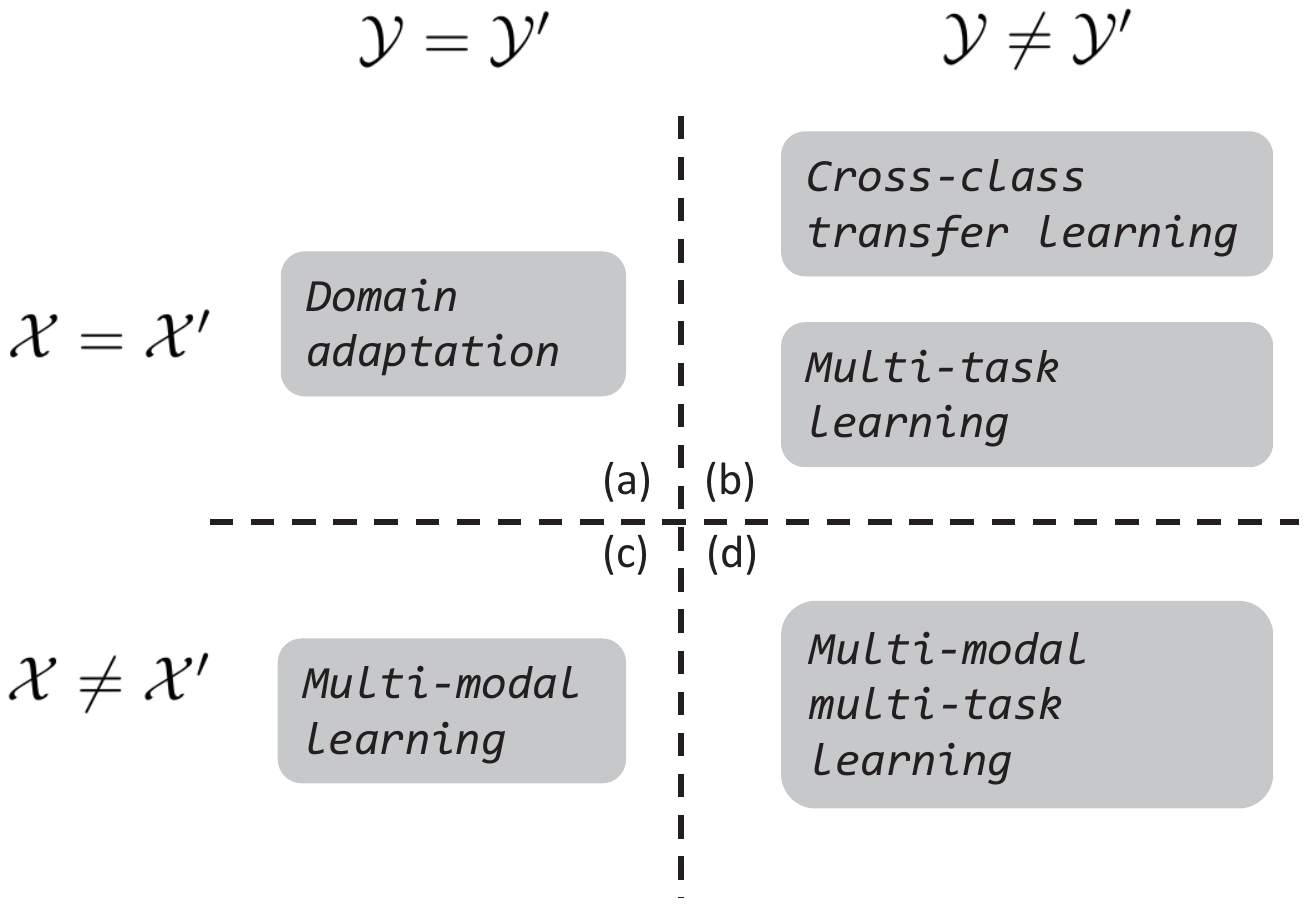}
\caption{Different tasks where the IID assumption no longer holds.}\label{fig:IID}
\end{figure}

\subsection{Learning with Interdependent Data}\label{sec:interdependent}
In traditional pattern recognition, the samples are assumed to be independent. However, in real world, we usually have some \emph{group} information (also denoted as \emph{set}, \emph{bag} or \emph{field} in the literature) for the samples, implying statistical dependencies among them. Let
\begin{equation}
\mathbf{X}_i = \{\boldsymbol{x}^i_1, \boldsymbol{x}^i_2, \ldots, \boldsymbol{x}^i_{n_i}\}, \ \mathbf{Y}_i = \{y^i_1, y^i_2, \ldots, y^i_{n_i}\}
\end{equation}
denote groups of samples and their corresponding labels. The purpose now is to learn the classifier and make decision with grouped data $\{\mathbf{X}_i, \mathbf{Y}_i\}_{i=1}^N$:
\begin{itemize}
\item In each group, the samples are no longer independent.
\item Different groups may not be identically distributed.
\item Different groups can have different cardinalities.
\end{itemize}

\subsubsection{Content consistency}
A straightforward and widely-considered case is that the samples in each group have the same label $y^i_1 = y^i_2 = \cdots = y^i_{n_i}, \forall i$, which is known as \emph{image set classification}~\cite{Hayat2015reconstruction} or \emph{group-based classification}~\cite{Samsudin2010group} in the literature. This kind of \emph{content consistency} in a group is very common in practice, for example, the temporal coherence between consecutive images in videos, the same object captured by multi-angle camera networks, classification based on long term observations~\cite{Shakhnaro2002long_term}, and so on. Each group $\mathbf{X}_i$ can be viewed as an unordered \emph{set} of samples, and therefore the task is to define the similarities between different sets, for example by: viewing each set as a \emph{linear subspace} and defining the similarity between two sets as canonical correlation~\cite{Kim2007cca}, describing each set with the Grassmann and Stiefel \emph{manifolds}~\cite{Turaga2011manifold} and using geodesic distances as metrics, representing each set as an \emph{affine hull}~\cite{Hu2012nearest_point} and calculating between-set distance from the sparse approximated nearest points, and so on. Besides viewing each set to lie on a certain geometric surface, deep learning framework based on minimum reconstruction error~\cite{Hayat2015reconstruction} can be used to automatically discover the underlying geometric structure for image set classification. The multiple samples in the same group (or set) will provide complementary information from different aspects like appearance variations, view-points, illumination changes, nonrigid deformations, and so on. Therefore, it offers new opportunities to improve the classification accuracy compared with single example based classification.

\subsubsection{Style consistency}
Besides content consistency, another situation is \emph{style consistency}~\cite{sarkar2005style}: the samples in a group are isogenous or generated by the same source. For example, in handwriting recognition, a group of characters produced by a certain writer are homogeneous with his/her individual writing style; in face recognition, face images can appear as different groups according to different poses or illumination conditions; in speech recognition, different speakers have different accents, and so on. These situations provide important group information, and in each group the style is consistent (other than content). Moreover, a new group unnecessarily enjoys the same style as the training groups, which means style transfer exists between training and test groups. In the literature, this problem is studied under the terminology of \emph{pattern field classification} by~\cite{sarkar2005style, veeramachaneni2005style, veeramachaneni2007analytical} where a \emph{field} is a group of isogenous patterns. Specifically, in~\cite{sarkar2005style} a class-style conditional mixture of Gaussians is used to model the isogenous patterns, in~\cite{veeramachaneni2005style} the dependencies among samples is modeled by second-order statistics with normally distributed styles, and in~\cite{veeramachaneni2007analytical} the intraclass and interclass styles are studied under adaptive classification. The traditional Bayes decision theory can also be extended to pattern field classification~\cite{zhang2011pattern}. By utilizing style consistency, classifying groups of patterns is shown to be much more accurate than classifying single patterns~\cite{zhang2011pattern} on various tasks like multi-pose face recognition, multi-speaker vowel classification, and multi-writer handwriting recognition.

\subsubsection{Group-level supervision}
A useful strategy to realize \emph{weakly-supervised learning} is that there is only \emph{group-level supervision} and the labels for the individual samples are not provided, which is a natural fit for numerous real-world applications and is denoted as \emph{multi-instance learning} (MIL)~\cite{Dietterich1997multi_instance} in the literature. A group of instances is denoted as a \emph{bag}, and although each bag has an associated label, the labels of the individual instances that conform the bag are not known. For example, in drug activity prediction~\cite{Dietterich1997multi_instance}, a molecule (bag) can adopt a wide range of shapes (instances) by rotating some of its internal bonds, and knowing a previously-synthesized molecule has desired drug effect does not directly provide information on the shapes. In image classification~\cite{Chen2006multi_instance}, a single image (bag) can be represented by a collection of regions, blocks or patches (instances), and the labels are only attached to images instead of the low-level segments.
The instances in each bag can be treated as non-IID samples~\cite{Zhou2009iid}, and not all of them are necessarily relevant~\cite{Amores2013multi_instance}: some instances may not convey any information about the bag class, or even come from other classes, thus providing confusing information. MIL usually deals with binary classification, and it is initially defined as~\cite{Dietterich1997multi_instance}: \emph{a bag is considered to be positive if and only if it contains at least one positive instance}. Relaxed and alternative definitions for MIL are presented by~\cite{Foulds2010multi_instance} to extend applications for different domains. Due to the property of weak supervision, MIL has found wide applications like image categorization~\cite{Chen2006multi_instance}, object localization~\cite{Cinbis2017multi_instance}, computer-aided diagnosis~\cite{Kandemir2015weak_diagnosis}, and so on.

\subsubsection{Decision making in context}
In above discussed situations, the \emph{order} of the samples in each group is actually ignored. However, the organizational structure of the samples gives us a very important information of \emph{context} which is historically shown to be crucial in pattern recognition~\cite{Haralick1983context}. For example, the \emph{linguistic context}~\cite{Suen1979ngram} takes place very naturally during the process of human reading. By using a language model~\cite{Bengio2003languagemodel}, the performance of many related tasks like speech recognition, character recognition and text processing can be significantly improved. Moreover, the spatial arrangement of the samples, known as \emph{geometric context}, is also an important piece of information for different pattern recognition tasks like~\cite{Felzenszwalb2010part} and~\cite{Wang2012hctr}. A widely-used strategy to learn from the context is viewing the samples as a \emph{sequence}, and many methods like hidden Markov models (HMMs)~\cite{Rabiner1989hmm}, conditional random fields (CRFs)~\cite{Lafferty2001crf} and recurrent neural networks (RNNs)~\cite{Sutskever2014sequence} can be used to model the dependencies among samples from the perspectives of Markov chain, conditional joint probability and long-short-term dependency respectively. Besides sequence, \emph{graph} is another useful representation for contextual learning, and recently graph neural networks~\cite{Scarselli2009graphnn, Kipf2017graph, Velickovic2018graphnn} have gained increasing popularity in various domains by modeling the dependencies between nodes in a graph via efficient and effective message passing among them. Moreover, other than structured input representation, the dependencies can also happen in the output space, known as \emph{structured output learning}~\cite{Tsochan2005structuredoutput}, which tries to predict more complex outputs such as trees, strings or lattices other than a group of independent labels. More details on this important issue can be found in~\cite{Nowozin2011structured}.

\subsubsection{Discussion}
By utilizing the dependencies among data, assigning labels for a group of patterns simultaneously will be more accurate and robust than labeling them separately. The key problem is how to define and learn from the group information. Content and style~\cite{Tenenbaum2000style_content} are two important factors in pattern recognition. The dependencies derived from content consistency and style consistency are useful information to improve the performance of group-based pattern recognition. Multi-instance learning which only requires group-level supervision is an effective strategy for weakly-supervised learning. The contextual information embedded in the order or arrangement of the samples is proved to be important for structured prediction. Besides using dependency to improve performance, automatic discovery of the relationship of samples (relational reasoning)~\cite{Santoro2017relational} is also an important direction.

\subsection{Domain Adaptation and Transfer Learning}\label{sec:domain_transfer}
As shown in Fig.~\ref{fig:IID}a, when both the feature space and label space are identical, the non-IIDness may happen on the conditional distribution. In this case, the \emph{domain adaptation} and \emph{transfer learning} are actually dealing with the same thing: there is usually a source domain with sufficient labeled data and a target domain with a small amount of labeled or unlabeled data, and the purpose is to reduce the distribution mismatch between two domains in supervised, unsupervised, or semi-supervised manners.

\subsubsection{Supervised fine-tuning}
When there exist some labeled data in target domain (\emph{supervised domain adaptation}), a simple and straightforward solution is to fine-tune the model on these extra labeled data. Actually, every pattern classifier which can be trained via \emph{incremental or online learning} can be used in this way for supervised domain adaptation. The model trained on the source domain can be viewed as not only a good initialization but also a regularization, and fine-tuning on target data will gradually reduce the distribution shift between two domains. Many classifiers can essentially be learned incrementally like neural networks trained with back-propagation. For non-incremental classifiers, we can also develop some counterpart algorithms for them, such as incremental decision tree~\cite{Utgoff1989incremental_tree}, incremental SVM~\cite{Laskov2006incrementalSVM}, and so on.

\begin{figure}[!t]
\centering
\includegraphics[width=0.9\columnwidth]{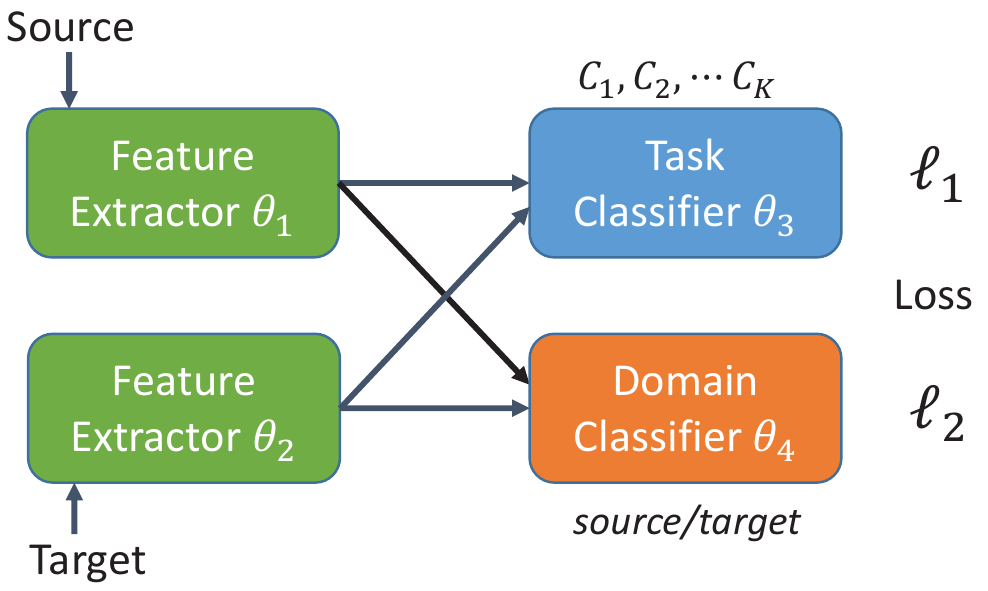}
\caption{An illustration of adversarial domain adaptation.}\label{fig:adv_domain}
\end{figure}

\subsubsection{Cross-domain mapping}
Another widely used strategy is learning cross-domain mappings to reduce the distribution shift. Let $\{\boldsymbol{x}^s,y^s\}$ and $\{\boldsymbol{x}^t, y^t\}$ denote source and target data, and $\theta$ denotes the parameters in classifier. The cross-domain mapping $\phi(\cdot)$ can be defined in various ways:
\begin{eqnarray}
\text{Parameter mapping:} \ p(y^s | \boldsymbol{x}^s; \theta) = p(y^t | \boldsymbol{x}^t; \phi(\theta)), \\
\text{Source mapping:} \ p(y^s | \phi(\boldsymbol{x}^s); \theta) = p(y^t | \boldsymbol{x}^t; \theta), \\
\text{Target mapping:} \ p(y^s | \boldsymbol{x}^s; \theta) = p(y^t | \phi(\boldsymbol{x}^t); \theta), \\
\text{Co-mapping:} \ p(y^s | \phi(\boldsymbol{x}^s); \theta) = p(y^t | \phi(\boldsymbol{x}^t); \theta).
\end{eqnarray}
In the first approach of \emph{parameter mapping}, the source and target distributions are matched using transformed parameters $\phi(\theta)$. For example, Leggetter and Woodland~\cite{leggetter1995speaker} use a linear transformation on the mean parameters of the hidden Markov model for speaker adaptation, and in~\cite{Rozantsev2018parametertransfer} a residual transformation network is used to map source parameters into target parameters. Another strategy is the \emph{source mapping} applied on the source data $\phi(\boldsymbol{x}^s)$~\cite{Courty2017transport}, and the transformed source data can be used together with the target data to train the classifier. Meanwhile, the \emph{target mapping} applies the mapping on target data $\phi(\boldsymbol{x}^t)$~\cite{zhang2013writer,Hoffman2013domain}, and the advantage is that the adaptation (learning of $\phi$) can happen after the training of the source classifier~\cite{zhang2013writer}. At last, we can also define the \emph{co-mapping} by projecting both the source and target data~\cite{Pan2011transferComponent} to a shared common space. The mapping $\phi$ can be either linear~\cite{leggetter1995speaker,zhang2013writer,Hoffman2013domain} or nonlinear~\cite{Rozantsev2018parametertransfer,Courty2017transport,Pan2011transferComponent}. Different criteria can be used to learn $\phi$, like maximum likelihood~\cite{leggetter1995speaker}, minimum earth mover distance~\cite{Courty2017transport}, minimum regularized Euclidean distance~\cite{zhang2013writer}, discriminative training~\cite{Hoffman2013domain,Rozantsev2018parametertransfer}, component analysis~\cite{Pan2011transferComponent}, and so on.

\subsubsection{Distribution matching}
The purpose of domain adaptation and transfer learning is to match the distributions of source and target domains. The \emph{importance re-weighting}~\cite{Shimodaira2000weighting} is a widely used strategy for distribution matching: each source sample is weighted by the importance factor $w(\boldsymbol{x}) = p_t(\boldsymbol{x}) / p_s(\boldsymbol{x})$ where $p_t(\boldsymbol{x})$ and $p_s(\boldsymbol{x})$ are target and source densities. Using weighted source samples to train the classifier will work well on target domain. However, density estimation is known to be a hard problem especially in high-dimensional spaces, therefore, directly estimating the importance without going through density estimation would be more promising as shown by~\cite{Huang2007sample_selection} and~\cite{Sugiyama2008importance}. Another problem in distribution matching is how to measure the discrepancy of two distributions. The \emph{maximum mean discrepancy} (MMD)~\cite{Gretton2007mmd} is a widely-used strategy by measuring the distance between the means of two distributions in a reproducing kernel Hilbert space (RKHS). It is shown~\cite{Smola2007hilbert} that MMD will asymptotically approach zero if and only if the two distributions are the same. Since MMD is easy to calculate and does not require the label information, it has been widely used as a regularization term for unsupervised domain adaptation~\cite{Long2015deepadaptation}. Other than only using the distance between first-order means as the measurement, Zhang et al.~\cite{Zhang2018covariance} propose aligning the second-order covariance matrices in RKHS for distribution matching. Besides MMD, many other kinds of distances, divergences, and information theoretical measurements can also be used for distribution matching, as discussed in the survey paper of~\cite{Zhang2017crossdataset}.

\subsubsection{Adversarial learning}
Recently, an increasingly popular idea of \emph{adversarial learning} tries to make the features from both domains to be as indistinguishable as possible. As shown in Fig.~\ref{fig:adv_domain}, the whole framework is composed of four components. The feature extractors $\theta_1$ (for source) and $\theta_2$ (for target) are usually defined as deep neural networks. The task classifier $\theta_3$ is used to perform the original $K$-way classification for both source and target data. Importantly, a domain classifier $\theta_4$ is used to judge whether a sample is from source or target domain (binary classification). Since we have two classifiers, here we can define two standard classification losses $\ell_1$ and $\ell_2$. The key of adversarial learning is that these losses are optimized like playing a min-max game:
\begin{eqnarray}
&&\theta_1 \Rightarrow \min \ell_1, \max \ell_2, \\
&&\theta_2 \Rightarrow \min \ell_1 \ (\text{optional}), \max \ell_2, \\
&&\theta_3 \Rightarrow \min \ell_1, \\
&&\theta_4 \Rightarrow \min \ell_2.
\end{eqnarray}
The purpose of $\min \ell_1$ is to guarantee classification accuracy while $\max \ell_2$ aims to confuse the domain classifier and make the feature distributions over two domains similar, thus resulting in domain-invariant features. To efficiently seek $\max_{\theta_1, \theta_2} \ell_2$, multiple strategies can be used, like using gradient reversal layer~\cite{Ganin2015gradient_reversal} to reverse the gradient in back-propagation, adopting inverted labels~\cite{Tzeng2017adv_domain} to calculate another surrogate fooled loss, and minimizing the cross-entropy loss against a uniform distribution~\cite{Tzeng2015uniform_distribution}. Moreover, the feature extractor modules can be designed as shared ($\theta_1=\theta_2$)~\cite{Ganin2015gradient_reversal}, partially shared~\cite{Rozantsev2019partially_shared}, or independent ($\theta_1 \neq \theta_2$)~\cite{Tzeng2017adv_domain}. Adversarial learning is efficient and effective for domain adaptation and transfer learning, and many subsequent improvements are still being gradually proposed~\cite{Wang2018deepdomainsurvey}.

\subsubsection{Multi-source problem}
In the above approaches, we assume that there is only a single source domain, but in practice, multiple sources~\cite{Mansour2009multisources} may exist during data collection, which is related to the style consistent pattern field classification problem in Section V-A.  The above discussed single-source methods can be extended accordingly to multi-source case. For example, the cross-domain mapping can be extended to multiple source-mappings with a shared Bayes classifier~\cite{zhang2011pattern}, the adversarial based method can be modified by replacing the binary domain classifier with a multi-way classifier representing multiple sources~\cite{Xu2018cocktail}, and so on.

\subsubsection{Discussion}
Domain adaptation and transfer learning are useful for many applications like speaker adaptation~\cite{leggetter1995speaker} in speech recognition, writer adaptation~\cite{zhang2013writer} in handwriting recognition, view adaptation~\cite{Kan2014face_domain} in face recognition, and so on. Fine-tuning is a straightforward and effective strategy for supervised adaptation, while cross-domain mapping is a general approach for supervised~\cite{Hoffman2013domain}, unsupervised~\cite{Courty2017transport}, and semi-supervised~\cite{zhang2013writer} adaptation. Traditional approaches usually focus on distribution matching, and adversarial learning has become the new trend for deep learning based adaptation.  Multi-source phenomenon is common in practice, and how to discover latent domains~\cite{Hoffman2012latentdomain} in mixed-multi-source data is an important and challenging problem.

\subsection{Multi-task Learning}\label{sec:multi_task}
As shown in Fig.~\ref{fig:IID}b, another case is that the feature spaces are identical but the label spaces are changed. For example, a face image can be classified into different races, ages, genders, and so on. These tasks are not independent, instead, they are complementary to each other, and learning one task is helpful for solving another. How to efficiently and effectively learn from multiple related tasks is known as \emph{multi-task learning}.

\begin{figure}[!t]
\centering
\includegraphics[width=\columnwidth]{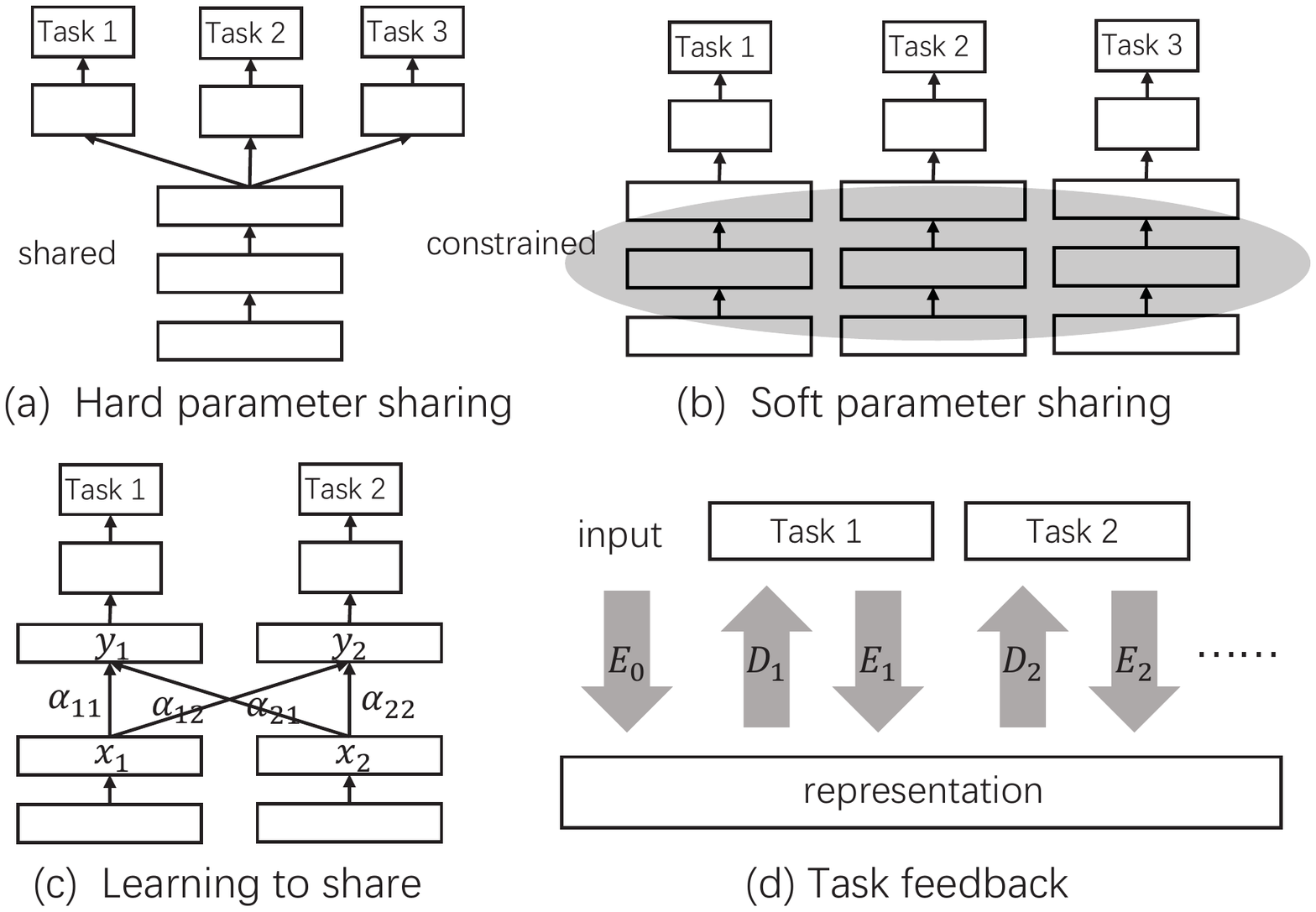}
\caption{An illustration of multi-task representation learning.}\label{fig:multi_task}
\end{figure}

\subsubsection{Transferable representation learning}
The first question for multi-task learning is that: can we find a generic feature representation that is transferable among different tasks? The traditional hand-crafted feature representations are usually task-specific, and new features need to be designed for new tasks. It has been shown by~\cite{Donahue2014decaf} that the features extracted from the deep neural network pre-trained with a large dataset are transferable to different tasks. The usual method is to train a base network, and then copy its first few layers to a target network, for which the remaining layers are randomly initialized and trained toward target task. There are multiple layers in deep neural networks, and the first layers usually learn \emph{low-level} features whereas the latter layers learn semantic or \emph{high-level} features. The low-level features are more general while the high-level features are more specific~\cite{Yosinski2014transferable}. Therefore, the transferabilities of features from bottom, middle, or top of a neural network are different, depending on the distance between the base task and target task~\cite{Azizpour2016transferability}: for similar tasks the later layers are more transferable, while for dissimilar tasks the earlier layers are preferred.

\subsubsection{Multi-task representation learning}
The second question for multi-task learning is that: can dealing with multiple tasks simultaneously be used to integrate different supervisory signals for learning an invariant representation? Since each task will produce a task-specific loss function, generally, as soon as you find yourself optimizing more than one loss function, you are effectively doing multi-task learning~\cite{Ruder2017overview_multi_task}. To learn multiple tasks jointly, there should be some \emph{shared} and \emph{task-specific} parameters in the architecture, and sharing what is learned during training different tasks in parallel is the central idea in multi-task learning. A straightforward approach is the \emph{hard parameter sharing}~\cite{Caruana1997multitask} (Fig.~\ref{fig:multi_task}a), which shares the bottom layers for all tasks and keeps several task-specific output layers. This is the most widely-used strategy in real applications due to its simplicity and effectiveness. However, it needs intensive experiments to find the optimal split position for shared and task-specific layers. Another approach is the \emph{soft parameter sharing} in Fig.~\ref{fig:multi_task}b, where each task has its own parameters which are regularized with some constraints to encourage similarity between them, like minimizing their $\ell_2$ distances~\cite{Duong2015parameter_sharing} or with some partially shared structure~\cite{Cao2018partial_share}. Other than using fixed sharing mechanism, another strategy is \emph{learning to share}. For example, the cross-stitch network~\cite{Misra2016cross_stitch} proposes to learn an optimal combination of shared and task-specific representations as shown in Fig.~\ref{fig:multi_task}c. The above approaches are based on sharing features among tasks, and the decision-making processes of each task are still independent. However, the solution of one task may be useful for solving other related tasks, indicating that we need \emph{task feedback} to update the representation as shown in Fig.~\ref{fig:multi_task}d. In this way, the performance of different tasks can be improved recurrently by utilizing the solutions (not only features) from other tasks~\cite{Bilen2016recurrent_multi_task}.

\subsubsection{Task relationship learning}
Finding the relationship between different tasks will make information sharing among tasks to be more selective and smooth. A straightforward strategy is using \emph{task clustering}~\cite{Thrun1996task_clustering} to partition multiple tasks into several clusters, and the tasks in the same cluster are assumed to be similar to each other. It is also possible to dynamically widen a thin neural network in a greedy manner to create a tree-like deep architecture for clustering similar tasks in the same branch as shown in~\cite{Lu2017tree_like_task}. Besides task clustering, many studies have also tried to learn both the per-task model parameters and the inter-task relationships simultaneously, where the task relationship can be formulated to be a matrix~\cite{Murugesan2016online_multi_task}, tensor~\cite{Almaev2015latent_task_structure}, or a nonlinear structure~\cite{Ciliberto2017nonlinear_output_relation}. This topic has attracted much attention in the research community, and the work of \emph{taskonomy}~\cite{Zamir2018taskonomy} has won the CVPR2018 best paper award. The taskonomy (task taxonomy) is a directed hyper-graph that captures the transferability among tasks: an edge between two tasks represents a feasible transfer, while the weight is the prediction of the transfer performance. With task relationship, the transfer learning performance would be improved, due to better transfer path from most related tasks to a target task, which can not only reduce the computational complexity of using all tasks but can also avoid the phenomenon of negative transfer caused by dissimilar tasks~\cite{Zamir2018taskonomy}.

\subsubsection{Discussion}
When multiple related tasks can be defined naturally, multi-task learning will significantly improve the performance for many problems like computer vision~\cite{Doersch2017multi_task} and natural language processing~\cite{Collobert2008multitask_nlp}. However, this requires labeled data for each task. A more efficient strategy is to use some \emph{auxiliary tasks} where data can be collected without manual labeling (to be discussed in Section~\ref{sec:self_supervised}). When multiple tasks are learned jointly, how to balance their loss functions becomes the key problem. A dominant approach is to assign each task a pre-defined weight. However, the optimal weight is expensive and time-consuming to find empirically. Therefore, Kendall et al.~\cite{Kendall2018weigh_losses} propose to learn the optimal weights automatically for multiple loss functions by considering the homoscedastic uncertainty of each task. Furthermore, gradient normalization~\cite{Chen2018gradnorm} can also be used to automatically balance training in deep multi-task models by dynamically tuning the gradient magnitudes.

\subsection{Multi-modal Learning}\label{sec:multi_modal}
The world surrounding us involves multiple modalities: we see objects, hear sounds, feel texture, smell odors, and taste flavors~\cite{Baltru2018multimodal}. Different from multi-task learning (Fig.~\ref{fig:IID}b) where multiple tasks are performed with the same input, the purpose of \emph{multi-modal learning} is to utilize the supplementary and complementary information in different modalities to complete a shared task (Fig.~\ref{fig:IID}c) or multiple related tasks (Fig.~\ref{fig:IID}d).

\subsubsection{Multi-modal representation and fusion}
In multi-modal learning, each instance can be viewed by multiple modalities, which can be fused at different levels. The first approach we can consider is \emph{signal-level fusion}, for example, the pansharpening of multi-resolution satellite images~\cite{Kang2014pansharpening}. However, different modalities usually have different data structures with different sizes or dimensions such as images, sound waves and texts, making them hard to be fused at raw-data level. Therefore, we must design some modality-wise representations, which could be either handcrafted features or learned with deep neural networks. For example, 2D/3D convolutional neural networks can be used for spatial structured signals like image, CT, fMRI, video, and so on, while recurrent neural networks can be used for temporal data like speech and text. With modality-wise feature extractions, different modalities are transformed into a unified space, and a straightforward fusion strategy is therefore \emph{feature-level fusion}, for example, the modality-wise representations can be concatenated into a longer representation or averaged (with learnable modality-wise weights) to a new representation~\cite{Liu2018combinemodalities}. After that, any traditional model can be learned on the fused feature representation for a given task. Another common strategy is \emph{decision-level fusion} which is widely-investigated in multiple classifier systems~\cite{Ho1994Multi_classifier_system}. This fusion strategy is often favored because different models are used on different modalities, making the system more flexible and robust to modality-missing as the predictions are made independently. Recently, due to the development of deep learning which learns hierarchical representations, the \emph{intermediate-level fusion}~\cite{Ramachandram2017multimodal_survey} is used to dynamically integrate modalities at different levels with an automatically learned and optimized fusion structure. Moreover, another important strategy is the \emph{learning-based fusion}, for example, using cross weights to gradually learn interactions of modalities~\cite{Rastegar2016cross_weights}, using multiple-kernel learning to learn optimal feature fusion~\cite{Gonen2011mkl}, learning sharable and specific features for different modalities~\cite{Wang2015multi_modal_sharable}, and so on.

A widely-occurring problem in multi-modal learning is modality-missing, i.e., some modalities are unaccessible for some instances during inference. The generative model such as deep Boltzmann machine~\cite{Srivastava2012deepBoltzmann} can be used to handle modality-missing by sampling the absent modality from the conditional distribution. We can also apply the modality-wise dropout~\cite{Neverova2016moddrop} during training to improve the generalization performance for modality-missing.

\subsubsection{Cross-modal matching, alignment, and generation}
Besides fusing multiple modalities to make accurate and robust predictions, another vibrant research direction causing increasing attentions is the \emph{cross-modal learning}~\cite{Baltru2018multimodal}. In this case, different modalities are embedded into a coordinated and well-aligned space, for example, the maximum correlated space by canonical correlation analysis (CCA)~\cite{Andrew2013deepCCA}, the semantic preserved space by joint embedding~\cite{Frome2013devise}, and so on. The embedded space enables \emph{cross-modal matching} tasks, such as retrieving the images that are relevant to a given textual query~\cite{Gu2018cross_modal_retrieval}, deciding which face image is the speaker given an audio clip of someone speaking~\cite{Nagrani2018cross_modal_matching}, and so on. A more difficult problem is \emph{cross-modal alignment} for finding correspondences between sub-items of instances from multiple modalities~\cite{Baltru2018multimodal}. For example, aligning the steps in a recipe to a video showing the dish being made~\cite{Malmaud2015cooking}, aligning a movie to the script or the book chapters~\cite{Zhu2015books_movies}. According to~\cite{Baltru2018multimodal}, the cross-modal alignment can be achieved either explicitly like using dynamic time warping and CCA, or implicitly like using the attention mechanism in deep neural networks. Another task \emph{cross-modal generation}, which seeks a mapping from one modality to another, has become very popular with an emphasis on language and vision~\cite{Baltru2018multimodal}, for example, generating the text description of an input image~\cite{Xu2015imagecaption}, inversely generating the image given a text description~\cite{Reed2016text_to_image}, and so on. The difficulty is increasing from matching, alignment, to generation, requiring much better understanding and high-level capture of the interaction and relationship between modalities.

\begin{figure}[!t]
\centering
\includegraphics[width=\columnwidth]{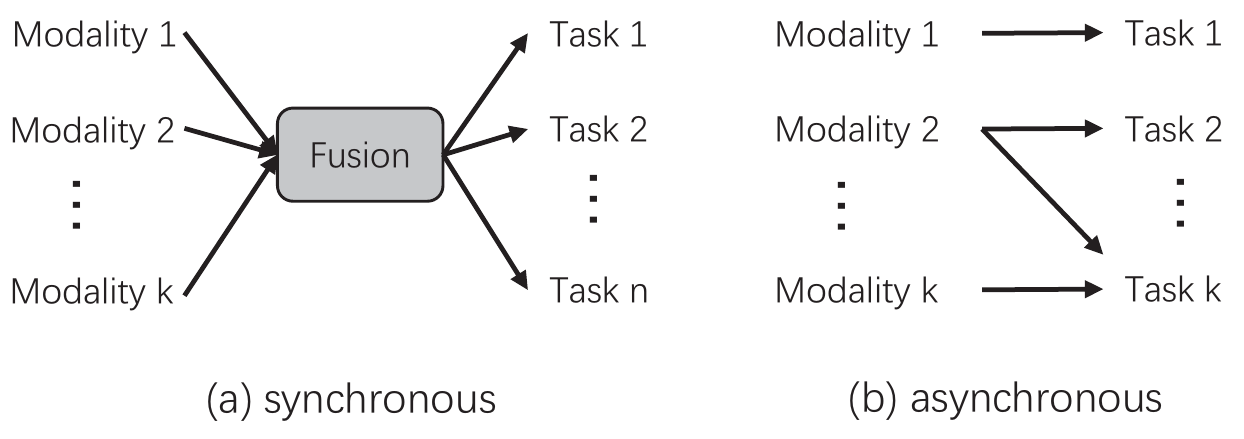}
\caption{Two types of multi-modal multi-task learning.}\label{fig:multi_modal_multi_task}
\end{figure}

\subsubsection{Multi-modal multi-task learning}
The last case we shall discuss is the multi-modal multi-task learning (Fig.~\ref{fig:IID}d), which can be partitioned into two types as shown in Fig.~\ref{fig:multi_modal_multi_task}. The first and also much simpler setting is the the \emph{synchronous} case, where all the modalities are available for solving each task. In this case, a fused representation of all modalities can be learned efficiently during joint training multiple related tasks, which is widely used in different applications such as disease diagnosis~\cite{Zhang2012multi_Modal_multi_task_disease}, traffic sign recognition~\cite{Lu2017traffic_sign}, autonomous driving~\cite{Chowdhuri2019autonomousdriving}, emotion recognition~\cite{Chen2017emotion_recognition}, and so on. As shown in Fig.~\ref{fig:multi_modal_multi_task}b, the second and also more challenging setting is the \emph{asynchronous} case, where different tasks may only rely on their own modalities. For example, image classification works on image, speech recognition deals with sound waves, while machine translation handles text data. Intuitively, it is hard to consider these problems jointly, since both their inputs and outputs are different. Moreover, it is also not clear about the common knowledge shared among these seemingly unrelated problems, and how much benefit we can get from combining them. An interesting work named ``\emph{one model to learn them all}''~\cite{Kaiser2017one_model_learn_all} shows us such possibility and potential, where a single deep neural network to learn multiple tasks from various modalities is designed using modality-specific encoders, I/O mixer, and task-specific decoders. The joint training of diverse tasks with asynchronous image, speech, and text modalities is shown to benefit from shared architecture and parameters. Amazingly, although seemed to be unrelated, incorporating image classification in training would help to improve the performance of language parsing, indicating that some computational primitives can be shared between different modalities and even unrelated tasks~\cite{Kaiser2017one_model_learn_all}.

\subsubsection{Discussion}
There are many practical problems that can benefit from multi-modal learning. In biometric applications, a person can be identified by face, fingerprint, iris, or voice. Although each modality is already distinguishable, their combination will improve both the accuracy and robustness. In autonomous driving, the fusion of multiple sensors (radar, camera, LIDAR, GPS, and so on) is necessary and important to make robust decisions. Multi-modal analysis will also make the decision making to be more explainable~\cite{Park2018multimodal_explan}. Moreover, the human brain is essentially both a multi-modal and a multi-task system: it continuously receives stimuli from various modes of the surrounding world and performs various perceptual and cognitive tasks. For a pattern recognition system, multi-modal perception increases the diversity on input while multi-task improves the diversity on output, and usually diversity will bring robustness. Therefore, joint multi-modal multi-task learning will be an inspiring and important future direction.

\section{Breaking Clean and Big Data Assumption}
Pattern recognition systems usually have strong abilities to memorize training data. As shown in~\cite{Zhang2017generalization}, even if we randomly change the labels of data completely, neural networks can still achieve near zero training error, indicating the strong capacity to fit training data. This is valuable if we have a clean (well-labeled) and large-enough (covering different variations) dataset, and fitting the training data in this case will usually also lead to good generalization performance. However, this assumption is hard to satisfy in real applications. Actually, \emph{clean data and big data are contradictory}: it is easy to collect a well-labeled small dataset, but it is impossible to manually label a big dataset without any error. Therefore, in order to improve the robustness on both the quality and quantity of data, first of all, the training process should be robust to noisy data, and second, particular learning strategies should be considered to reduce the dependence on large amounts of data. To reach this goal, we present discussions and summarizations from the following four perspectives.

\subsection{Supervised Learning with Noisy Data}\label{sec:noisy_data}
In supervised learning, the noises in data can be partitioned into three types: (1) \emph{label noise}: the sample is valid but the label is wrong due to mislabeling; (2) \emph{sample noise} (or attribute noise): the sample is noisy but the label is valid, for example, samples caused by corruption, occlusion, distortion, and so on; (3) \emph{outlier noise}: both the sample and label are invalid, for example, samples from a new not-care class or a totally noisy signal, but still labeled as one of the classes to be classified. To deal with noisy data, different approaches have been proposed in the literature. Frenay and Verleysen~\cite{Frenay2014noise_survey} have surveyed many methods for \emph{label noise} before the year of 2014. Complementarily, we consider all the three noise types and focus more on methods developed in recent years.

\subsubsection{Robust loss}
The unbounded loss function will usually over-emphasize the noisy data, and hence, the decision boundaries will deviate severely from the optimal one. Therefore, the first solution for learning with noisy data is to redefine the loss function to be bounded. The convex functions are usually unbounded, and therefore, most redefined robust loss functions are non-convex~\cite{Shirazi2009nonconvex}. For example, the ramp loss~\cite{Brooks2011ramp_loss} and truncated hinge loss~\cite{Wu2007truncated} set an upper bound on hinge loss by allowing a maximum error for each training observation, resulting in a non-convex but robust SVM~\cite{Ertekin2011nonconvex}. The correntropy-induced loss~\cite{Xu2018robust} with properties of bounded, smooth, and non-convex is shown to be robust when combined with kernel classifiers. For deep neural networks, as suggested by~\cite{Ghosh2017noise}, the categorical cross entropy loss is sensitive to label noise, and a comparison of different loss functions tolerant to label noise is given in~\cite{Ghosh2015noise}.

\subsubsection{Noise transition}
For a sample $\boldsymbol{x}$ with annotation $y$ (either correct or wrong), we use $\hat{y}$ to denote its ground-truth (clean) label. Now, the labeling noises can be modeled probabilistically by $p(y|\hat{y}, \boldsymbol{x})$ which is usually a complex process. However, we can assume $p(y|\hat{y}, \boldsymbol{x}) = p(y|\hat{y})$: noisy label depends only on true label and not on the sample. This is an approximation of real-world labeling process and can still be useful in some certain scenarios~\cite{Patrini2017noise}, for example, there are usually some confusable (similar) categories which are hard for human labelers to distinguish, regardless of the specific samples. In this case, we can simply use a \emph{noise transition matrix} $T$ to specify the probability of one label being wrongly annotated to another $T_{ij} = p(y=j | \hat{y}=i)$. Since $p(y|\boldsymbol{x}) = \sum_{\hat{y}} p(\hat{y}|\boldsymbol{x})p(y|\hat{y})$, we can modify the loss function to be~\cite{Sukhbaatar2015noise}:
\begin{equation}
\mathcal{L} = -\frac{1}{n} \sum_{i=1}^n \log \left( \sum_{j=1}^{k} p(j|\boldsymbol{x}_i)  T_{y_i, j} \right),
\end{equation}
where $k$ is the number of classes. The matrix $T$ can be estimated from data and subsequently fixed during classifier training~\cite{Patrini2017noise}, jointly estimated with the classifier~\cite{Sukhbaatar2015noise,Jindal2016noise}, or estimated with human-assistance~\cite{Han2018noise}. Moreover, Vahdat~\cite{Vahdat2017noise} proposes an undirected graphical model to directly model $p(y|\hat{y}, \boldsymbol{x})$ other than $p(y|\hat{y})$.

\subsubsection{Cleaning}
Another approach is explicitly detecting and removing the noisy data. An effective strategy is using ensemble learning to filter noisy data~\cite{Brodley1999mislabel}.  In ensemble learning, different classifiers are complementary to each other, hence, examples which are in contradiction with most learners can be identified confidently as noisy. With this kind of approach, the \emph{data pruning} method~\cite{Angelova2005pruning} is shown to significantly improve generalization performance. For large-scale dataset cleaning, the \emph{partitioning filter}~\cite{Zhu2003noise} is proposed for noise identification from large distributed datasets. Another approach is directly incorporating the noise detection into the objective function of the learning machine, for example, the robust SVM approach~\cite{Xu2006robust} uses a binary indicator variable for each sample to explicitly mark it as noisy or clean. In this way, the noisy data can be automatically suppressed and no loss is charged for them during training. Similar idea is also used for learning distance metric from the noisy side information~\cite{Huang2010robust_metric}.

\begin{table}[!t]
\renewcommand{\arraystretch}{1.6}
\caption{Methods on supervised learning with noisy data.}\label{table:noisy}
\centering
\begin{tabular}{|c|c|c|c|}
\hline
&Label Noise &Sample Noise &Outlier Noise \\ \hline
Robust Loss &\Checkmark &\Checkmark &\Checkmark \\ \hline
Noise Transition &\Checkmark &\XSolid &\XSolid \\ \hline
Cleaning &\Checkmark &\Checkmark &\Checkmark \\ \hline
Reweighting &\Checkmark &\Checkmark &\Checkmark \\ \hline
Relabeling &\Checkmark &\XSolid &\XSolid \\ \hline
\end{tabular}
\end{table}

\subsubsection{Reweighting}
Reweighting is a soft-version of cleaning: assigning small weights for noisy data other than completely removing them~\cite{Liu2016noisy}. The work of~\cite{Wang2018iterative} proposes a \emph{reweighting module} by a Siamese network to distinguish clean labels and noisy labels under iterative learning. Moreover, the \emph{cleanNet}~\cite{Lee2018cleannet} assigns weights as the sample-to-label relevance calculated from a joint neural embedding network for measuring the similarity between a sample and its noisy labeled class. The work of \emph{mentorNet}~\cite{Jiang2018mentornet} treats the base model as the \emph{studentNet} and a mentorNet is used to provide a curriculum (the reweighting scheme) for studentNet to focus on samples with probably correct labels, which is shown to significantly improve the performance on real-world large-scale noisy dataset of WebVision.

\subsubsection{Relabeling}
We can also correct the labels of noisy data by \emph{relabeling} in the learning process. In~\cite{Xiao2015noisy}, a probabilistic graphical model is used to simulate the relationship between samples, labels and noises, for deducing the true label with an EM-like algorithm. The \emph{bootstrapping}~\cite{Reed2015bootstrap} is also used for relabeling the noisy data by updating the labels as convex combination of original noisy label and current prediction of classifier iteratively. Similarly, Tanaka et al.~\cite{Tanaka2018joint} propose to learn model parameters and true labels alternatively under a joint optimization framework.

\subsubsection{Discussion}
As shown in Table~\ref{table:noisy}, different methods can handle different data noises. Although the technical details are different, the purposes of robust loss, cleaning, and reweighting are actually similar, i.e., reducing the influence of noisy data in learning process. Therefore, they can be used for all the three noise types. The noise transition and relabeling methods are only suitable for the case of label noise, however, they are efficient and effective strategies to improve robustness when the noises in data are mainly caused by mislabeling.

\subsection{Unsupervised (Self-supervised) Learning}\label{sec:self_supervised}
In traditional pattern recognition~\cite{jain2000review}, unsupervised learning is usually referred to \emph{data clustering}, however, nowadays, more emphasis is actually placed on \emph{unsupervised representation learning}~\cite{Bengio2013Review}, where good and transferrable feature representation is learned from large amounts of unlabeled data. A widely-used strategy is \emph{self-supervised learning}, a specific instance of supervised learning where the targets are directly generated from the data and therefore no need for labeling.

\subsubsection{Reconstruction-based}
Since the data are unlabeled, a straightforward strategy is to use them as both the inputs and targets to learn a compressed representation with an encoder $f(\boldsymbol{x})$ and decoder $g(\boldsymbol{x})$ to minimize the reconstruction error:
\begin{equation}
\min_{f,g} \sum_{\boldsymbol{x}} \left\| g \left( f(\boldsymbol{x}) \right) - \boldsymbol{x} \right \|.
\end{equation}
The first approach following this idea is principal component analysis (PCA)~\cite{fukunaga1990introduction} which learns a linear subspace via $f(\boldsymbol{x}) = W \boldsymbol{x}$ and $g(\boldsymbol{x})=W^{\top} \boldsymbol{x}$ with an orthogonal matrix $W$ for projection. The restricted Boltzmann machine~\cite{hinton2006science} and auto-encoder~\cite{Vincent2008autoencoder} are also reconstruction based methods and can be viewed as nonlinear extensions of PCA. From then on, various improvements have been proposed such as denoising auto-encoder~\cite{Vincent2008autoencoder}, contractive auto-encoder~\cite{Rifai2011contractive}, variational auto-encoder~\cite{Kingma2013variational}, and so on. The \emph{split-brain auto-encoder}~\cite{Zhang2017split} splits the model into two disjoint sub-networks for cross-channel prediction, which transforms the \emph{reconstruction} objective to a \emph{prediction} based one, making the learned feature representation more semantic and meaningful.

\subsubsection{Pseudo label with clustering}
We can also assign some \emph{pseudo labels} to the data, and then transform the problem to a supervised learning task. For unlabeled data, a natural idea is to use some clustering algorithm to partition the data into different clusters, and then the cluster identities can be viewed as the pseudo label to learn representations. However, a challenge is that the clustering relies heavily on a good representation, and conversely the learning of representation also requires good clustering results as supervision, resulting in a chicken-or-egg-first problem. To solve this problem, the alternative learning strategy~\cite{Bautista2016cliqueCNN, Yang2016joint} can be used for joint unsupervised learning of the representations and clusters.

\subsubsection{Pseudo label with exemplar learning}
Other than clustering, another method of \emph{exemplar learning} views each sample as a particular class, the pseudo label now is the sample identity, and the purpose is to separate all training samples from each other as much as possible. The \emph{exemplar-CNN}~\cite{Dosovitskiy2016exemplar} treats each patch (with random transformations) in an unlabeled image as a particular class, and the classifier is trained to separate all these classes. In this way, the learned representation not only ensures that different patches can be distinguished but also enforces invariance to specified transformations. Another approach uses \emph{noise as targets}~\cite{Bojanowski2017noise} to learn the representation and the one-to-one matching of training samples to uniformly sampled vectors for separating every training instance. In exemplar learning, each instance is treated as a distinct class of its own, therefore, the number of classes is the size of the entire training set. The computational challenges imposed by large number of classes need to be carefully considered~\cite{Wu2018instance} in exemplar learning. Recently, the \emph{momentum contrast} (MoCo)~\cite{He2019MoCo} proposes using a dictionary as a queue and a momentum update mechanism to efficiently and effectively realize the idea of exemplar learning, and shows that the gap between unsupervised and supervised representation learning can be closed in many vision tasks.

\subsubsection{Surrogate tasks for computer vision}
\begin{figure}[!t]
\centering
\includegraphics[width=0.95\columnwidth]{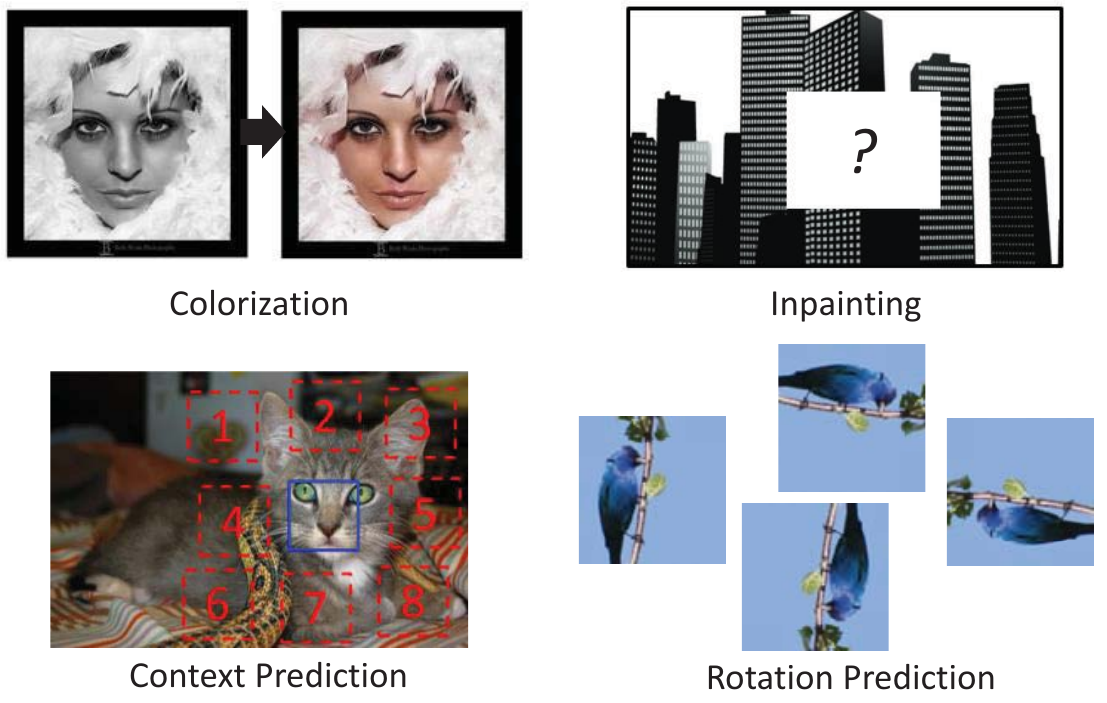}
\caption{Different types of self-supervision for visual tasks.}\label{fig:self_supervised}
\end{figure}

Recently, another interesting trend for unsupervised learning is seeking help from some \emph{surrogate tasks} for which the labels or targets come for ``\emph{free}'' with the data. For example, as shown in Fig.~\ref{fig:self_supervised}, learning to colorize grayscale image~\cite{Zhang2016color,Larsson2016color}, learning by inpainting~\cite{Pathak2016inpaint} (to generate the contents of an missing area in image conditioned on its surroundings), learning by context prediction~\cite{Doersch2015context} (to predict the position of one patch relative to another patch in image), learning by solving jigsaw puzzles~\cite{Noroozi2016puzzle} (geometric rearrangement of random-permutated patches), learning by predicting image rotations~\cite{Gidaris2018rotation}, and so on. For all these tasks, the supervisory signal can be easily obtained automatically, and therefore, there is no need to worry about the insufficiency and labeling of the training data.

Although these tasks seem to be simply defined, doing well on these tasks requires the model to learn meaningful and semantic representations. For example, for inpainting the model needs to understand the content of the image to produce a plausible hypothesis for the missing part, in context prediction the model should learn to recognize objects and also their parts to predict relative positions. Therefore, the learned representations from these surrogate tasks can transfer well to more complicated tasks. Similar approaches can also be found on video related tasks~\cite{Wang2015video, Arandjelovic2017look}.

\subsubsection{Surrogate tasks for natural language}
The idea of using surrogate tasks to learn representations from unlabeled data can also be used for other tasks like natural language processing, such as the \emph{language model} for predicting what comes next in a sequence~\cite{Dai2015semi_sequence, Radford2018pre_training}. The work of BERT~\cite{Devlin2018bert} proposes two novel surrogate tasks for unsupervised learning. The first is \emph{masked language model} which randomly masks some of the tokens from the input and the objective is to predict the original vocabulary identity of the masked word based on its context. The second task is \emph{next sentence prediction} which is a binary classification task to predict whether a sentence is next to another sentence or not. These strategies have achieved new benchmark performance on eleven language tasks~\cite{Devlin2018bert}, indicating unsupervised pre-training has become an important integral part for language understanding.

\subsubsection{Discussion}
Due to the abundantly available unlabeled data, unsupervised or self-supervised learning is a long pursued objective for representation learning. In addition to the methods discussed above, it is hopeful to see more and more effective and interesting self-supervised methods in the future. Besides classification, self-supervised learning can also be used for regression problems~\cite{Liu2019self_regression}. Since different approaches have been proposed from different aspects, the combination of multiple self-supervised methods through multi-task learning~\cite{Doersch2017multi_task} (Section~\ref{sec:multi_task}) is an inspiring direction.

\subsection{Semi-supervised Learning}\label{sec:semi_supervised}
Semi-supervised learning (SSL) deals with a small number of labeled data and a large amount of unlabeled data simultaneously, and therefore, can be viewed as a combination of supervised and unsupervised learning. In the literature, a wide variety of methods have been proposed for SSL, and comprehensive surveys can be found in~\cite{Chapelle2006semi_supervised} and~\cite{zhu2006semi_survey}. Nowadays, new progress especially deep learning based approaches have become the new state-of-the-art, therefore, in this section we focus on recent advances in SSL.

\subsubsection{Reconstruction-based}
A straightforward strategy for SSL is combining the supervised loss on labeled data and unsupervised loss on unlabeled data to build a new objective function. The \emph{ladder network}~\cite{Rasmus2015ladder} combines the denoising auto-encoder (as unsupervised learning for every layer) with supervised learning at the top layer for SSL. The \emph{stacked what-where auto-encoder}~\cite{Zhao2016what} uses a convolutional net for encoding and a deconvolutional net for decoding to simultaneously minimize a combination of supervised and reconstruction losses. Similarly, Zhang et al.~\cite{Zhang2016semi_large} take a segment of the classification network as encoder and use mirrored architecture as decoding pathway to build several auto-encoders for SSL.

\subsubsection{Self/co/tri-training}
Another approach is using initial classifier to predict \emph{pseudo labels} for unlabeled data and then retraining classifier with all data. This process is repeated iteratively for boosting both the accuracy of pseudo labels and the performance of classifier. Following this, the first idea is \emph{self-training}~\cite{Lee2013pseudo_label} where a single model is used to predict the pseudo label as the class with maximum predicted probability, which is equivalent to the \emph{entropy minimization}~\cite{Grandvalet2005entropy} in SSL. The \emph{co-training}~\cite{Blum1998co_training} uses two different models to label unlabeled data for each other. Moreover, the \emph{tri-training}~\cite{Zhou2005tri-training} utilizes bootstrap sampling to get three different training sets for building three different models. For example, in the \emph{tri-net}~\cite{Chen2018trinet} approach, three different modules are learned and if two modules agree on the prediction of the unlabeled sample confidently, the two modules will teach the third module on this sample. The strategies of self/co/tri-training are efficient to implement and also effective for SSL.

\subsubsection{Generative model}
Generative model is another widely used strategy for SSL. For example, the Gaussian mixture model~\cite{zhu2006semi_survey} can be used to maximize the joint likelihood of both labeled and unlabeled data using EM algorithm. The \emph{variational auto-encoder} (VAE)~\cite{Kingma2014deep_generative} can be used for SSL by treating the labels as additional latent variables. A recent trend is using \emph{generative adversarial network} (GAN)~\cite{Goodfellow2014gan} for SSL by setting up an adversarial game between a discriminator $D$ and generator $G$. In original GAN, $D$ is a binary classifier. To apply GAN for SSL, $D$ is modified to be a $k$-class model~\cite{Springenberg2016catgan} or extended to $k+1$ classes ($k$ real classes and one fake class)~\cite{Salimans2016gans}. For labeled data, $D$ should minimize their supervised loss. For unlabeled data, $D$ is then trained to minimize their uncertainty (e.g. by entropy minimization) to $k$ classes. Moreover, $D$ should also try to distinguish the generated samples by either maximizing their entropy (uncertainty) to $k$ classes~\cite{Springenberg2016catgan} or classifying them to the additional fake class~\cite{Salimans2016gans}. Meanwhile, $G$ is trained from the opposite direction to generate realistic samples for the $k$ classes. By using GAN for SSL, the advantages are two-fold. First, it can generate synthetic samples for different classes which serve as additional training data. Second, even bad examples~\cite{Dai2017badgan} from the generator will benefit SSL, because they are lying in low-density areas of the manifold which will guide the classifier to better locate decision boundary.

\subsubsection{Perturbation-based}
Most deep learning models utilize randomness to improve generalization. Therefore, multiple passes of an individual sample through the network might lead to different predictions, and the inconsistency between them can be used as the loss for unlabeled data. Let $f$ be a model with parameter $\theta$, and $\eta, \eta'$ as different randomness (data augmentation, dropout, and so on). The mean squared error is used by~\cite{Sajjadi2016perturbation} to minimize the inconsistency of the predictions $\| f(\boldsymbol{x}|\theta, \eta) - f(\boldsymbol{x}|\theta, \eta') \|_2^2$. This is denoted as $\Pi$-model in~\cite{Laine2017temporal}, where each sample is evaluated twice and the difference of predictions is minimized.

Actually, this can also be explained from the \emph{teacher-student} viewpoint. For each unlabeled sample, a teacher $\mathcal{T}(\boldsymbol{x})$ is used to guide the learning of the student $f$ by minimizing
\begin{equation}
\| f(\boldsymbol{x}| \theta, \eta) - \mathcal{T}(\boldsymbol{x}) \|_2^2.
\end{equation}
In $\Pi$-model the teacher is another evaluation with a different perturbation $\mathcal{T}(\boldsymbol{x}) = f(\boldsymbol{x}|\theta, \eta')$. However, a single evaluation can be very noisy, therefore, the \emph{temporal ensembling}~\cite{Laine2017temporal} proposes the use of exponentially moving average (EMA) of the predictions to form a teacher:
\begin{equation}
\mathcal{T}(\boldsymbol{x}) \leftarrow \alpha \mathcal{T}(\boldsymbol{x}) + (1-\alpha) f(\boldsymbol{x}|\theta, \eta'),
\end{equation}
where $0<\alpha<1$ is a momentum, and each unlabeled sample has a teacher that is the temporal ensembling of previous predictions. Moreover, the \emph{mean teacher}~\cite{Tarvainen2017meanteacher} proposes to average model parameters other than predictions, i.e., the teacher uses EMA parameters $\theta'$ of student model $\theta$:
\begin{equation}
\theta' \leftarrow \alpha \theta' + (1-\alpha) \theta,
\end{equation}
and now the teacher is $\mathcal{T}(\boldsymbol{x}) = f(\boldsymbol{x}|\theta', \eta')$, which is the same model with student but using historically-averaged parameters. The perturbation-based approaches will smooth the predictions on unlabeled data. Moreover, other than random perturbations, the \emph{virtual adversarial training}~\cite{Miyato2018virtual} can be used to find a worst-case perturbation for better SSL.

\subsubsection{Global consistency}
The perturbation-based approach is actually seeking a kind of \emph{local consistency}: samples that are close in input space (due to perturbation) should also be close in output space. However, a more important idea is \emph{global consistency}: samples forming an underlying structure should have similar predictions~\cite{Zhou2004local_global}.
To better utilize global consistency, the traditional graph-based SSL is combined with deep learning~\cite{Kamnitsas2018ssl}, by dynamically creating a graph in the latent space in each iteration batch to model data manifold, and then regularizing with the manifold for a more favorable state of class separation. Another interesting approach of \emph{learning by association}~\cite{Haeusser2017association} considers global consistency from a different perspective: imagine a walker from labeled data to unlabeled data according to the similarity calculated from latent representation, and then the walker will go back from unlabeled data to labeled data. Correct walks that start and end at the same class are encouraged, and wrong walks that end at a different class are penalized. The cycle-consistent association from labeled data to unlabeled ones and back can be efficiently modeled using transition probabilities~\cite{Haeusser2017association}, and therefore is an effective strategy to pursue global consistency in SSL.

\subsubsection{Discussion}
In SSL, the massive unlabeled data and the scarce labeled data reveal the underlying manifold of the entire dataset, and by letting the predictions for all samples to be smooth on the manifold, more accurate decision boundary can be obtained compared to purely supervised learning. Reconstruction-based methods can learn a better representation from unlabeled data, while pseudo labels can be used for iterative training on unlabeled data. Recently, a new trend is using GAN to model the distribution of labeled and unlabeled data for SSL. Moreover, perturbation-based methods utilize the randomness in deep neural network to seek local consistency on unlabeled data, and how to effectively consider global consistency in deep learning based SSL still needs more exploration.

\subsection{Few-shot and Zero-shot Learning}\label{sec:few_shot}
\begin{figure}[!t]
\centering
\includegraphics[width=\columnwidth]{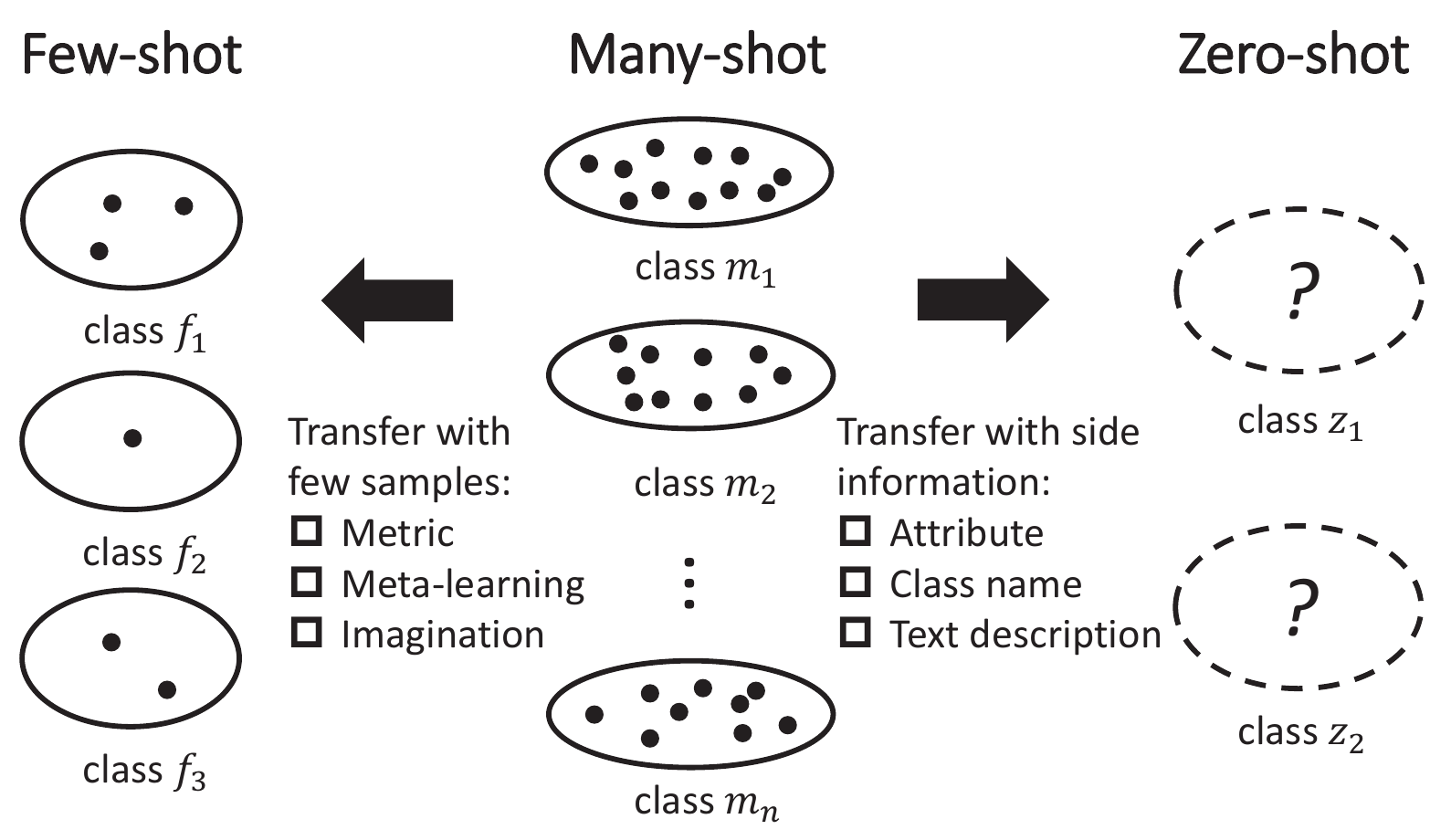}
\caption{From many-shot to few-shot and zero-shot learning.}\label{fig:few_shot}
\end{figure}

In human intelligence, we can instantly learn a novel concept by observing only a few examples from a particular class. However, the state-of-the-art approaches in machine intelligence are usually highly data-hungry. This difference has inspired an important research topic of \emph{few-shot learning} (FSL)~\cite{Lake2015human}. In FSL, we have a many-shot dataset $\mathcal{M}$ and few-shot dataset $\mathcal{F}$ (usually \emph{$k$-shot $n$-way}: $k$ labeled samples for each of the $n$ classes, and $k$ is small like 1 or 5). The samples in $\mathcal{M}$ have disjoint label space with $\mathcal{F}$. The purpose of FSL is to extract transferrable knowledge from $\mathcal{M}$ to help us perform better on $\mathcal{F}$, as illustrated in Fig.~\ref{fig:few_shot}.

\subsubsection{Metric learning}
Under the principle that test and training conditions must match, the \emph{episode} based training~\cite{Vinyals2016matching} is used to mimic the $k$-shot $n$-way setting. Specifically, in each training iteration, an \emph{episode} is formed by randomly selecting $n$ classes with $k$ samples per class from $\mathcal{M}$ to act as the \emph{support} set $\mathcal{S}=\{(\boldsymbol{x}_i,y_i)\}_{i=1}^{kn}$, and meanwhile a fraction of the remainder samples from those $n$ classes are selected as the \emph{query} set. This support/query split is designed to simulate the real application situation on $\mathcal{F}$. The purpose now is to define the probability $P(y|\boldsymbol{x},\mathcal{S})$. Although the underlying classes are different between $\mathcal{M}$ and $\mathcal{F}$, the learned $P(y|\boldsymbol{x},\mathcal{S})$ is hoped to be transferable between them, which can be viewed as a point-to-set metric. For example, in~\cite{Vinyals2016matching}, a deep neural network embedded space is learned to calculate such a metric. To learn a better embedding (or metric), a memory module can be used~\cite{Cai2018memory_match} to explicitly encode the whole support set $\mathcal{S}$ into memory for defining $P$. Moreover, different criteria can be used to learn the metric like the mean square error~\cite{Sung2018relation_network} and the ranking loss~\cite{Triantafillou2017few}. Since each support set $\mathcal{S}$ is designed to be few-shot, the metric learned in this way can be transferred to unseen categories which also have few examples.

\subsubsection{Learning to learn (meta-learning)}
The \emph{learning to learn} or \emph{meta-learning} is to train another learner at a higher level for guiding the original learner. In~\cite{Wang2016learn_to_learn}, the meta-learner is defined as the transformation from the model parameters learned from few samples to the model parameters learned from large enough samples, which can be viewed as a \emph{model-model regression}. Another approach~\cite{Kozerawski2018one_shot} instead uses the \emph{sample-model regression} as the meta-learner by transforming each single sample directly to the classifier parameters. Denote original model as $\varphi(\boldsymbol{x}; W)$ where $W$ represents the parameters to be learned, instead of learning $W$ directly, a meta-learner $\omega(\boldsymbol{x}; W')$ is used in~\cite{Bertinetto2016one_shot} to map $\boldsymbol{x}$ to $W$. Now, the model becomes $\varphi(\boldsymbol{x}; \omega(\boldsymbol{x}; W'))$ and the task is changed to learn $W'$. The meta-learner $\omega(\boldsymbol{x}; W')$ can be used to predict any parameters of another network $\varphi$ like linear layers or convolutional layers~\cite{Bertinetto2016one_shot}, and once learned, the parameters for any novel category can be predicted by a simple forward pass. Besides predicting parameters, other meta-learning methods such as learning the optimization algorithm~\cite{Ravi2017optimization} or initialization~\cite{Finn2017meta_learning} can also be used for FSL. Since the tasks considered in meta-learning is category-agnostic, they can be well transferred to new few-shot categories.

\begin{table*}[!t]
\renewcommand{\arraystretch}{1.6}
\caption{Evaluation metric and representative performance for different robustness issues in pattern recognition.}\label{table:performance}
\centering
\begin{tabular}{|c|m{0.4\textwidth}|m{0.4\textwidth}|}

\multicolumn{1}{c}{\cellcolor{mygray}\textbf{Open-world}} &\multicolumn{1}{c}{\cellcolor{mygray}\textbf{Evaluation Metric}} &\multicolumn{1}{c}{\cellcolor{mygray}\textbf{Representative Performance}} \\ \hline
Section~\ref{sec:kk} &\emph{\textbf{Classification accuracy}}: the ratio of number of correctly classified patterns to the total number of patterns, evaluated on a test dataset different from the training dataset. &On a benchmark 1000-class ImageNet~\cite{imagenet} dataset, human-level accuracy is 94.9\% (top-5), and ResNet~\cite{he2016resnet} could achieve 96.43\% accuracy. On a smaller 10-class MNIST~\cite{lecun1998gradient} dataset, it is common to achieve more than 99\% accuracy (top-1). \\ \hline
Section~\ref{sec:outlier} &\emph{\textbf{Rejection performance}}: a threshold is used to distinguish normal and abnormal patterns, and to evaluate overall performance, threshold-independent metric is used like area under curve. &To detect \emph{not}MNIST from MNIST~\cite{Hendrycks2017baseline}, the AUROC (area under receiver operating characteristic curve) is 85\% and AUPR (area under precision-recall curve) is 86\%. \\ \hline
Section~\ref{sec:uk} &\emph{\textbf{Adversarial robustness}}: Let $\triangle(\boldsymbol{x}, f) = \min_{\eta} \| \eta \|_2$ subject to $f(\boldsymbol{x}+\eta) \neq f(\boldsymbol{x})$, the robustness of classifier $f$~\cite{Dezfooli2016deepfool} is $\rho(f) = \mathds{E}_{\boldsymbol{x}} \frac{\triangle(\boldsymbol{x},f)}{\|\boldsymbol{x}\|_2}$ where $\mathds{E}_{\boldsymbol{x}}$ is expectation over data. &On ILSVRC the adversarial robustness~\cite{Dezfooli2016deepfool} is $2.7 \times 10^{-3}$ for CaffeNet and $1.9 \times 10^{-3}$ for GoogLeNet, indicating that: a perturbation with $1/1000$ magnitude as the original image is sufficient to fool state-of-the-art deep neural networks. \\ \hline
Section~\ref{sec:uu} &\emph{\textbf{Class-incremental capacity}}: the changing trend of classification accuracy as the number of classes increased. &On ILSVRC as the number of classes incremented from 100 to 1000, the accuracy is reduced from about 90\% to 45\%~\cite{Rebuffi2017incremental}. \\ \hline

\multicolumn{1}{c}{\cellcolor{mygray}\textbf{Non-iid}} &\multicolumn{1}{c}{\cellcolor{mygray}\textbf{Evaluation Metric}} &\multicolumn{1}{c}{\cellcolor{mygray}\textbf{Representative Performance}} \\ \hline
Section~\ref{sec:interdependent} &\emph{\textbf{Contextual learning ability}}: the performance improvement caused by learning from a group of interdependent patterns.  &By integrating geometric and linguistic contexts,~\cite{Wang2012hctr} improved the correct rate of handwritten Chinese text recognition from 69\% to 91\%.\\ \hline
Section~\ref{sec:domain_transfer} &\emph{\textbf{Adaptability and transferability}}: the performance of adaptation and transfer between different (i.e., source and target) domains which have different data distributions. &Through writer adaptation with style transfer mapping,~\cite{zhang2013writer} achieved more than 30\% error reduction rate on a large scale handwriting recognition database CASIA-OLHWDB. \\ \hline
Section~\ref{sec:multi_task} &\emph{\textbf{Multi-task cooperation ability}}: the gain of considering multiple tasks simultaneously compared to handling them independently. &The taskonomy~\cite{Zamir2018taskonomy} reduced the number of labeled samples needed for solving 10 tasks by roughly $\frac{2}{3}$ (compared to training independently) while keeping the performance nearly the same. \\ \hline
Section~\ref{sec:multi_modal} &\emph{\textbf{Multi-modal fusion ability}}: the boost of performance by utilizing the complementary information in different modalities. &By fusing multiple modalities at several spatial and temporal scales,~\cite{Neverova2016moddrop} won the first place out of 17 teams on the ChaLearn 2014 ``looking at people challenge gesture recognition track''. \\ \hline

\multicolumn{1}{c}{\cellcolor{mygray}\textbf{Noisy Small Data}} &\multicolumn{1}{c}{\cellcolor{mygray}\textbf{Evaluation Metric}} &\multicolumn{1}{c}{\cellcolor{mygray}\textbf{Representative Performance}} \\ \hline
Section~\ref{sec:noisy_data} &\emph{\textbf{Noisy data tolerance}}: the stability of classification performance when the training data contain a certain percentage of noises. &On CIFAR10, with clean data the accuracy is 93\%, when 30\% of the data are noisy, the accuracy is deteriorated to 72\%, and a noise-robust model~\cite{Tanaka2018joint} can recover the accuracy to 91\%.\\ \hline
Section~\ref{sec:self_supervised} &\emph{\textbf{Self-supervised capability}}: performance of learning from purely unlabeled data under some self-supervised mechanisms. &Using instance discrimination as the self-supervision, the unsupervised MoCo~\cite{He2019MoCo} outperformed its supervised pre-training counterpart in 7 vision tasks on many datasets. \\ \hline
Section~\ref{sec:semi_supervised} &\emph{\textbf{Semi-supervised capability}}: performance of joint learning from massive unlabeled data and scarce labeled data. &On ImageNet, the accuracy of supervised learning (100\% labeled data) is 96\%, a semi-supervised model~\cite{Tarvainen2017meanteacher} (10\% labeled and 90\% unlabeled data) could achieve 91\% accuracy. \\ \hline
Section~\ref{sec:few_shot} &\emph{\textbf{Few-shot generalization}}: knowledge transfer ability from learning of old classes to new classes with few or even zero data. &On \emph{mini}ImageNet~\cite{Snell2017prototypical} with 5-class, the 1-shot (one sample per class) accuracy is 49\% and the 5-shot accuracy is 68\%. On CUB~\cite{Snell2017prototypical} with 50-class, the 0-shot (no sample but side information of attribute is available) accuracy is 55\%. \\ \hline
\end{tabular}
\end{table*}

\subsubsection{Learning with imagination}
Another explanation for the FSL ability of humans is that we can easily visualize or imagine what novel objects should look like from different views although we only see very few examples. This has inspired the \emph{learning with imagination} to produce additional training examples. As pointed out by~\cite{Hariharan2017hallucinating}, the challenge of FSL is that the few examples can only capture very little of the category's intra-class variation. To solve this, we can use the many-shot dataset $\mathcal{M}$ to learn the intra-class transformations of samples and then augment the samples in $\mathcal{F}$ along these transformations~\cite{Hariharan2017hallucinating}. In another work~\cite{Wang2018imaginary}, a \emph{hallucinator} is trained by taking a single example of a category and producing other examples to expand the training set. The hallucinator is trained jointly with the classification model, and the goal is to help the algorithm to learn a better classifier~\cite{Wang2018imaginary}, which is different from other data generation model like GAN~\cite{Goodfellow2014gan} whose goal is to generate realistic examples. Actually, the effectiveness of learning with imagination comes from the recovery of the intra-class variation missed in FSL.

\subsubsection{Zero-shot learning}
An extreme case of FSL is \emph{zero-shot learning} (ZSL)~\cite{Xian2019zeroshot} where there is no example for the novel categories. In this case, some \emph{side information} is needed to transfer the knowledge from previously learned categories to novel categories (Fig.~\ref{fig:few_shot}), including attributes~\cite{Lampert2014attribute}, class names~\cite{Halah2016miss_zero}, word vector~\cite{Frome2013devise}, text description~\cite{Qiao2016zero_textual}, and so on. In attribute-based ZSL~\cite{Lampert2014attribute}, attributes are typically nameable properties that are present or not for a certain category. In this way, multiple binary attribute-specific classifiers can be trained independently. After that, for a new class, training samples are no longer required, we only need the attribute associations for this class, and a test sample can be effectively classified by checking its predicted attributes. Besides user-defined attributes, learning the latent attributes~\cite{Peng2018latent_attribute} and the class-attribute associations~\cite{Halah2016miss_zero} can further improve performance. A more general approach for ZSL suitable for different side information is embedding based approach~\cite{Akata2016label_embedding}, where two embedding networks are learned for both samples and side information, and the similarity between them are measured using Euclidean, cosine, or manifold distance~\cite{Fu2018zero_graph}. In the embedded space, nearest neighbor search (cross-modal match) can then be efficiently used for ZSL. The last approach for ZSL is synthesizing class-specific samples conditioned on their side information for unseen classes~\cite{Long2018zero_synthesis}, which can be implemented in various ways like: data generation at feature-level~\cite{Xian2018featre_generating} or sample-level~\cite{Verma2018zero_synthesis}, using variational auto-encoder~\cite{Verma2018zero_synthesis} or generative adversarial network~\cite{Zhu2018zero_shot_gan}, conditioned on attributes~\cite{Long2018zero_synthesis} or text descriptions~\cite{Zhu2018zero_shot_gan}, and so on.

\subsubsection{Discussion}
Building a good model totally from scratch with a small number of observations is difficult, and actually, the FSL abilities of human beings are based on our abundant prior experiences of dealing with related tasks. Therefore, as pointed out by~\cite{Feifei2006one_shot}: the key insight for FSL is that the categories we have already learned can give us information that helps us to learn new categories with fewer examples. Therefore, FSL can be viewed as cross-class transfer learning. Moreover, humans are good at ZSL because we have other knowledge sources (like book and Internet) from which we can infer what a new category looks like. Therefore, ZSL is more like cross-modal learning (Section~\ref{sec:multi_modal}). Although many approaches have been proposed in the literature, few-shot and zero-shot learning are still urgently needed skills for machine intelligence.

\section{Concluding Remarks}
This paper considers the robustness of pattern recognition from the perspective of three basic assumptions, which are reasonable in controlled laboratory environments for pursuing high accuracies, however, will become unstable and unreliable in real-life applications. To improve robustness, we present a comprehensive literature review of the approaches trying to break these assumptions:
\begin{itemize}
\item For breaking \emph{closed-world assumption}, we partition the open-space into four components: the known known corresponding to the empirical risk, the known unknown corresponding to the outlier risk, the unknown known corresponding to the adversarial risk, and the unknown unknown corresponding to the open class risk.
\item For breaking \emph{independent and identically distributed assumption}, we first discuss the problems in learning with interdependent data, then review recent advances in domain adaptation and transfer learning, and finally analyse the multi-task and multi-modal learning for increasing the diversity on both output and input of the system.
\item For breaking \emph{clean and big data assumption}, we first introduce supervised learning with noisy data, then review un/self-supervised and semi-supervised learning to learn from unlabeled data and surrogate supervision, and lastly discuss few-shot and zero-shot learning to transfer knowledge from big-data to small-data.
\end{itemize}

With the above approaches, we can improve the robustness of a pattern recognition system by: growing continuously with \emph{changing concepts} in open world, adapting smoothly with \emph{changing environments} under non-identical conditions, and learning stably with \emph{changing resources} under different data quality and quantity. Actually, these are fundamental issues in robust pattern recognition, because these changing factors will usually greatly affect the stability of final performance in practice. Furthermore, in continuous use of a pattern recognition system, other than being a static model, how to make it evolvable during \emph{lifelong learning}~\cite{Pentina2015lifelong, Parisi2019lifelong} or \emph{never-ending learning}~\cite{Mitchell2018never_ending} is an important step towards real intelligence. Through breaking the three basic assumptions, we can actually eliminate the main obstacles in reaching this goal.

Unlike the traditional closed-world classification which is usually evaluated by accuracy, how to evaluate the recognition performance in open and changing environments is a big issue. Besides accuracy, other evaluation measurements reflecting the ability in dealing with the changing factors are more important. As shown in Table~\ref{table:performance}, when considering many other evaluation metrics (different from the classification accuracy), it is obvious that pattern recognition is far from solved. Moreover, in the research community, different tasks are usually evaluated with different metrics and databases. How to build a general benchmark for evaluating the robustness by integrating different metrics together is an important future task for pattern recognition.

Different from the widely-used empirical risk minimization, theoretical analysis to unify different open-world risks will become the foundation for future classifier design. A future pattern recognition system should acquire complementary information from interdependent data in different modalities and boost itself through the cooperation of multiple tasks by adaptively learning from a few labeled, unlabeled or noisy data. Although many attempts have been proposed in the literature, most of them try to solve a single problem from a single perspective. However, the three basic assumptions are actually related, and through joint consideration many new research problems can be raised, such as open-world domain adaptation~\cite{Busto2017opendomain}, open-world semi-supervised learning~\cite{Oliver2008semi}, cross-modal domain adaptation~\cite{Hoffman2016cross_modal_adaptation}, multi-task self-supervised learning~\cite{Doersch2017multi_task}, few-shot domain adaptation~\cite{Motiian2017few_domain}, and so on. Future research of a unified framework to deal with the open-world, non-i.i.d., noisy and small data issues simultaneously is the ultimate goal of robust pattern recognition.

Besides the robustness issues, many other problems are also important for pattern recognition. For example, the \emph{interpretability}~\cite{molnar2019interpretable} of the model: besides high accuracy, the system also needs to explain why such a prediction is made, for increasing our confidence and safety in trusting the result. Some traditional classifiers like decision tree and logistic regression are interpretable, but how to make other models especially black-box deep neural networks explainable~\cite{Zhou2019interpreting} is an important task. Another important issue is \emph{computational efficiency}~\cite{Han2016compression}. Besides big data, strong computing power is also a key for the success of modern pattern recognition technologies. In order to widen the application scope and also reduce resource consumption, the compression and acceleration of pattern recognition models are of great values for practical applications. Since pattern recognition can be viewed as the simulation of human brain perception ability which enables machine to recognize objects or events in sensing data, how to effectively learn from neuroscience for developing \emph{brain-inspired}~\cite{Poo2016china_brain}, \emph{biologically-plausible}~\cite{Bengio2015biologically} or \emph{psychophysics-driven}~\cite{Richard2019psychophysics} pattern recognition models is an inspiring future direction. With more attentions and efforts paid to these important issues in pattern recognition, the gap between human intelligence and machine intelligence can be narrowed in the foreseeable future.

\begin{appendices}
\section{Background References}
\begin{itemize}
\item Pattern Recognition~\cite{nagy1968state}~\cite{Fu1980pr}~\cite{jain2000review}~\cite{fukunaga1990introduction}~\cite{Duda2012book}~\cite{Bishop2006book}
\item Deep Learning~\cite{lecun2015nature}~\cite{goodfellow2016deep}~\cite{schmidhuber2015deep}
\item Outlier Detection~\cite{Hodge2004outlier}
\item Adversarial Example~\cite{Fawzi2017robust}
\item Open Set Recognition~\cite{Scheirer2013openSet} 
\item Class-incremental Learning~\cite{Rebuffi2017incremental}
\item Contextual Learning~\cite{Haralick1983context}
\item Domain Adaptation~\cite{Daume2006domainadaptation}
\item Transfer Learning~\cite{Pan2010transfer}
\item Multi-task Learning~\cite{Caruana1997multitask}
\item Multi-modal Learning~\cite{Baltru2018multimodal}
\item Learning with Noise~\cite{Frenay2014noise_survey}
\item Representation Learning~\cite{Bengio2013Review}
\item Self-supervised Learning~\cite{Kolesnikov2019Self_supervised}
\item Semi-supervised Learning~\cite{Chapelle2006semi_supervised}
\item Few-shot Learning~\cite{Feifei2006one_shot}
\item Zero-shot Learning~\cite{Xian2019zeroshot}
\end{itemize}
\end{appendices}

\bibliographystyle{plain}
\bibliography{robust}

\end{document}